    \RequirePackage{fix-cm}
    \documentclass[twocolumn]{svjour3}          
    \smartqed  
    \usepackage{setspace}
\usepackage{capt-of}
\usepackage{floatrow}
\usepackage{cite}
\usepackage{natbib}
\usepackage{graphicx}
\usepackage{float}
\usepackage[normalem]{ulem}
\usepackage{subfigure}
\usepackage{amsfonts}
\usepackage{ifthen}
\usepackage{booktabs}
\usepackage{amssymb}
\usepackage{supertabular}
\usepackage{array}
\usepackage{multirow}
\usepackage{fancyhdr}
\usepackage{fancyheadings}
\usepackage{psfrag}
\usepackage{amsmath}
\usepackage{hhline}
\usepackage{type1cm}
\usepackage{lettrine}
\usepackage{cite}
\usepackage{fancybox}
\usepackage{enumitem}
\usepackage{bm}
\usepackage{pifont}
\usepackage{threeparttable}
\usepackage{soul}

\usepackage{tikz}
\usetikzlibrary{shapes.geometric, arrows}
\usepackage[linesnumbered,ruled,vlined]{algorithm2e}
\usepackage[noend]{algpseudocode}
\usepackage{bchart}
\usepackage{mathtools}

\usepackage{pifont}

\usepackage{xcolor}
\usepackage{color, colortbl}
\definecolor{citecolor}{HTML}{0071bc}
\definecolor{tabhighlight}{HTML}{e5e5e5}
\usepackage[colorlinks,citecolor=citecolor]{hyperref}

\definecolor{amethyst}{rgb}{0.6, 0.4, 0.8}
    
    \makeatletter
    \renewcommand\paragraph{
      \@startsection{paragraph} 
      {4} 
      {\z@} 
      {.5em \@plus1ex \@minus.2ex} 
      {-.5em} 
      {\normalfont\normalsize\bfseries} 
    }
    \makeatother

\usepackage{amsmath,amsfonts,bm}

\def\eqref#1{equation~\ref{#1}}

\def\1{\bm{1}}

\def\vb{{\bm{b}}}
\def\vc{{\bm{c}}}
\def\vd{{\bm{d}}}

\def\vf{{\bm{f}}}

\def\vo{{\bm{o}}}

\def\vr{{\bm{r}}}

\def\vx{{\bm{x}}}

\def\vz{{\bm{z}}}

\def\mW{{\bm{W}}}

\DeclareMathAlphabet{\mathsfit}{\encodingdefault}{\sfdefault}{m}{sl}
\SetMathAlphabet{\mathsfit}{bold}{\encodingdefault}{\sfdefault}{bx}{n}

\def\gL{{\mathcal{L}}}
\def\gM{{\mathcal{M}}}

\def\gP{{\mathcal{P}}}

\def\gR{{\mathcal{R}}}

\def\gW{{\mathcal{W}}}

\DeclarePairedDelimiter\norm{\lVert}{\rVert}%
    \renewcommand{\vec}[1]{\boldsymbol{#1}}

\newcommand{\nerf}{\mathcal{N}}

\newcommand{\point}{\vec{x}}
\newcommand{\tpoint}{\vec{x}_{t}}
\newcommand{\spoint}{\vec{x}_{s}}
\newcommand{\pointtemplate}{\vec{x}_{0}}

\newcommand{\code}{\vec{z}}

\def\ECCVMETHODNAME{\textit{DDF}}
\newcommand{\NeRF}{{\mathcal{N}}}

\newcommand{\eg}{\textit{e.g.}}
\newcommand{\ie}{\textit{i.e.}}

\newlength\savewidth

    \journalname{IJCV}
    
\begin{document}
    \sloppy
    \title{Correspondence Distillation from NeRF-based GAN}
    \author{Yushi Lan         \and
            Chen Change Loy  \and
            Bo Dai
    }
    \institute{Yushi Lan \at
                  S-Lab, Nanyang Technological University, Singapore \\
                  \email{yushi001@e.ntu.edu.sg}           
               \and
               Chen Change Loy \at
                  S-Lab, Nanyang Technological University, Singapore \\
                  \email{ccloy@ntu.edu.sg}
               \and
               Bo Dai \at
                  Shanghai AI Laboratory \\
                  \email{doubledaibo@gmail.com}
    }
    
    \date{Received: date / Accepted: date}

    \maketitle

    \newcommand*{\DDFPath}{DDF}
\begin{abstract}
    \label{sec:abstract}
    The neural radiance field (NeRF) has shown promising results in preserving the fine details of objects and scenes. However, unlike mesh-based representations, it remains an open problem to build dense correspondences across different NeRFs of the same category, 
    which is essential in many downstream tasks.
    The main difficulties of this problem lie in the implicit nature of NeRF and the lack of ground-truth correspondence annotations.
    In this paper, we show it is possible to bypass these challenges by leveraging the rich semantics and structural priors encapsulated in a pre-trained NeRF-based GAN.
    Specifically,
    we exploit such priors from three aspects,
    namely 
    1) a dual deformation field that takes latent codes as global structural indicators,
    2) a learning objective that regards generator features as geometric-aware local descriptors,
    and
    3) a source of infinite object-specific NeRF samples.
    Our experiments demonstrate that such priors lead to 3D dense correspondence that is accurate, smooth, and robust.
    We also show that established dense correspondence across NeRFs can effectively enable many NeRF-based downstream applications such as texture transfer.
    \end{abstract}

    \begin{figure}[H] 
    \centering
    \noindent\includegraphics[width=1.\linewidth]{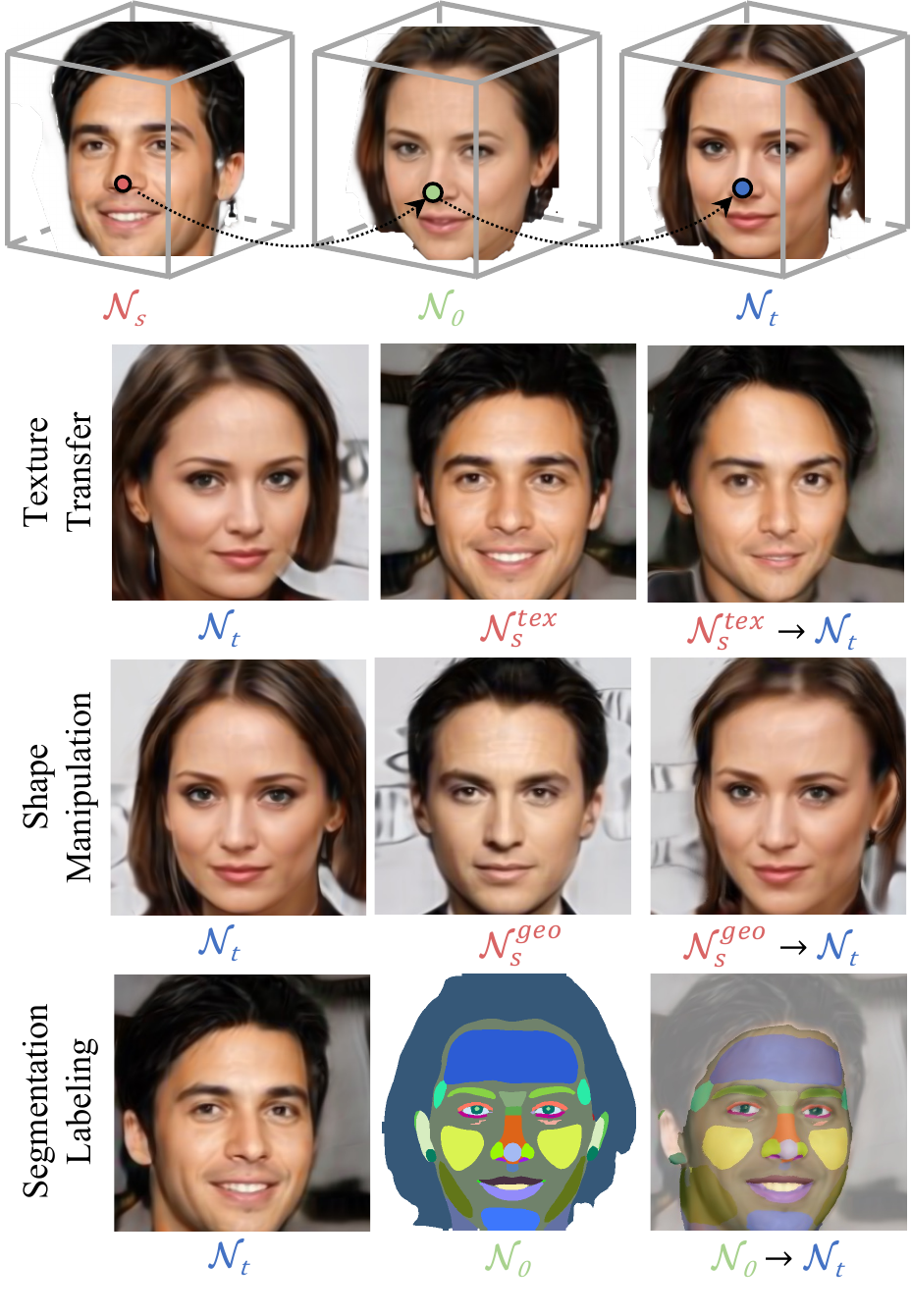}
    \centering
    \vskip -0.75cm
    \caption{\textbf{Dense correspondence across two objects represented as NeRF.} 
    We propose Dual Deformation Field (\ECCVMETHODNAME{}) to establish 3D dense correspondences between two objects represented as NeRF.
    This is achieved by a composite deformation (row 1):
    given a source-target NeRF pair $\nerf_s$ and $\nerf_t$, which we aim to establish correspondences on,
    for a point $\spoint$ from the source NeRF $\nerf_s$,
    we first establish its correspondence with a point on the template NeRF $\nerf_0$,
    which is further deformed to the target NeRF $\nerf_t$.
    Our training is fully self-supervised and could facilitate a series of downstream applications,
    such as texture transfer (row 2),
    shape manipulation (row 3) and
    1-shot view-consistent segmentation transfer (row 4). 
    }
    \label{fig:teaser}
    \end{figure}

\section{Introduction}
\label{sec:intro}


The success of neural radiance fields (NeRF) \citep{mildenhall2020nerf} 
has led to remarkable progress in learning 3D representations.
Unlike voxel- and mesh-based methods, NeRF represents each 3D object as a distribution of coordinate-based volume densities and view-dependent colors.
And by approximating this distribution with a continuous parametric function, 
NeRF shows great potential in capturing geometric scene details and rendering realistic novel views.

In this work, we study the potential of \emph{establishing dense correspondence across two objects represented as NeRF},
which is an important prerequisite for many downstream applications such as 
{texture transfer, manipulation and segmentation transfer, 
as shown in the Fig.~\ref{fig:teaser}.}
%
This task is non-trivial.
First, existing methods for building dense correspondence across two objects mainly focus on mesh-based representations. It is infeasible to directly apply and adapt them to NeRF.
Unlike meshes that have explicit vertices and surfaces,
NeRF lacks an explicit surface,
preventing us from resorting derivatives of neural fields~\citep{yang2021geometry} as the shape surface descriptors.
Moreover, existing methods~\citep{Litany2017DeepFM} often require ground-truth correspondence annotations in training,
which are hard to obtain for NeRF-based object representations.

\begin{figure}[t] 
\centering
\noindent\includegraphics[width=1.\linewidth]{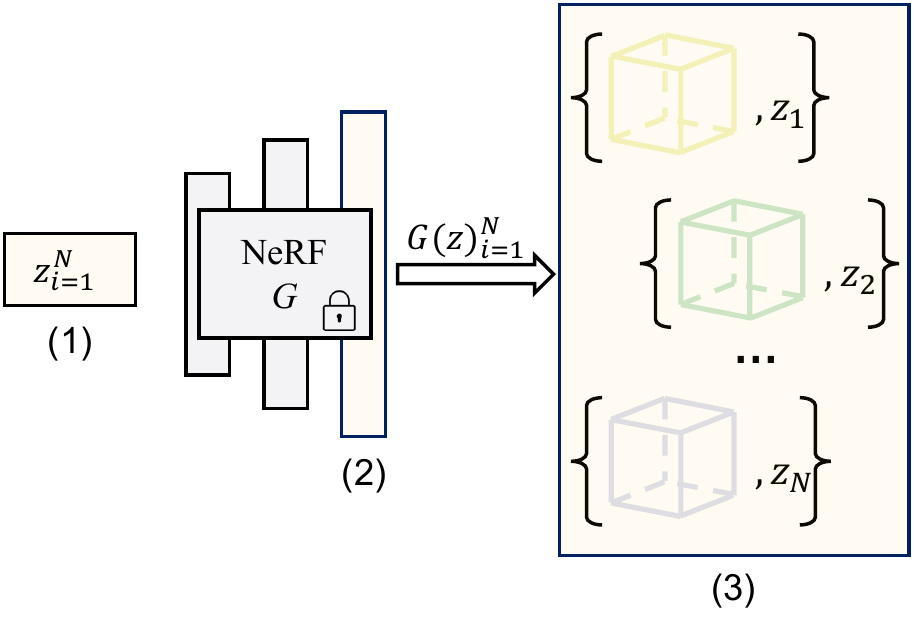}
\centering
\caption{\textbf{The triple role of a NeRF-based GAN}: 
We retrofit a pretrained NeRF-based GAN into triple roles: 
(1) the latent codes $z_{i=1}^{N}$ serve as holistic structure descriptors; 
(2) the extracted generator features serve as geometry-aware local descriptors;
and (3) the sampling space of pretrained $G$ could serve as an infinite object-specific dataset.
}
\label{fig:teaser2}
\end{figure}

To overcome the aforementioned limitations, we present a novel approach that exploits NeRF-based generative adversarial networks (GANs)~\citep{Chan2021piGANPI,GIRAFFE,Schwarz2020NEURIPS} to facilitate the learning of dense correspondence in NeRF.
Specifically, NeRF-based GANs treat image synthesis as novel views rendering from its intermediate NeRF representation.
%
Our key idea is to employ its generator, $G$, to play a \textbf{triple role}, as shown in Fig.~\ref{fig:teaser2}:

\noindent
\textbf{1)}
Since the generator of a GAN is a latent variable model
that learns a mapping $z \rightarrow G(z)$,
the associated latent code $z$ shall capture the underlying structure of the generated object NeRF $G(z)$ in a pretrained GAN.
Therefore,
this latent code naturally serves as a holistic global structure descriptor for building conditional models
that generalize to different object NeRFs of a category of interest.

\noindent
\textbf{2)}
As a representation learning architecture, $G$ can serve as a robust semantic embedding function that maps corresponding coordinates across different NeRFs into semantically similar features. 
%
Based on such cross-instance feature similarity,
we can thus naturally use features by $G$ as geometric-aware local descriptors. 
%
%

\noindent
\textbf{3)}
$G$ can also serve as a source of infinite object-specific NeRFs $\mathcal N_{i=1}^{\inf}$ for training,
where it is flexible to adjust the complexity of sampled NeRFs through the latent codes.

We name our approach as \textbf{Dual Deformation Field} (\ECCVMETHODNAME{}). 
While our \ECCVMETHODNAME{} does not limit the choice of NeRF-based GAN, we use $\pi$-GAN~\citep{Chan2021piGANPI} due to its simplicity and promising synthesis results.
Our adaptation of the pre-trained $\pi$-GAN starts with considering its first role in model construction,
where we treat latent codes of $\pi$-GAN as additional conditions.
Specifically,
we regard dense correspondence between NeRFs as a coordinate-based deformation field from the source NeRF to the target NeRF.
Instead of learning a single deformation field conditioned on a pair of source and target latent codes,
we use a fixed template NeRF as the bridge 
and learn two separate deformation fields,
namely a backward deformation field $B$ and a forward deformation field $F$.
In our formulation,
$B$ always treats the template NeRF as the target, taking only the source latent code as input.
Similarly,
$F$ always treats the template NeRF as the source, taking the target latent code as the condition.
Such a decomposition substantially alleviates the learning complexity. In addition, 
the dense correspondence between any two NeRFs can be easily established by combining $F$ and $B$. 

Benefit from the second role of $\pi$-GAN,
\ECCVMETHODNAME{} can learn without ground-truth correspondence annotations.
Specifically,
for any coordinate in the source NeRF,
we can obtain its corresponding coordinate in the target NeRF from \ECCVMETHODNAME{}.
Since the features of $\pi$-GAN are geometric-aware descriptors,
we, therefore, compute generator features for these estimated corresponding coordinates
and apply feature-wise cosine similarity as the primary learning objective.

Finally,
as $\pi$-GAN provides infinite object-specific NeRF samples for training, 
in practice we further control the complexity of sampled NeRFs
by 
mixing the latent codes of sampled NeRFs with that of the template NeRF,
we can use training samples of low deformation complexity in the beginning,
and gradually move to samples with higher complexity as training proceeds.
We found this strategy improves the training-time efficiency and stability.
%

This is an early attempt that establishes the dense correspondence between two NeRF-based object representations.
Without any ground-truth correspondence annotations,
dense correspondence is established by mining rich semantic and structural priors from a pre-trained NeRF-based GAN.
In the challenging category of human faces, the proposed method produces high-quality dense correspondences with promising robustness and generality.
Various tasks such as texture transfer and segmentation transfer are tested to demonstrate the potential of our method in downstream tasks. Code and models will be released. 

\section{Related Work}
\label{sec:related_work}
\noindent
\textbf{3D Shape Correspondences.} 
The problem of establishing dense correspondences between 3D shapes 
is of key importance to a series of downstream tasks \citep{loper2015smpl,Egger20203DMF},
and has been studied extensively in recent survey~\citep{Kaick2011ASO,Sahillioglu2019RecentAI}.
Traditional approaches build correspondence between shapes represented by mesh or point clouds. 
They can be roughly divided into registration-based and similarity-based methods,
where the former adopts Laplacian coordinate $\delta_i$ for vertex $v_i$ as geometric preservation descriptor after registration.
Similarity-based solutions do not change the geometry of given shapes and calculate the similarity between vertices with learnable feature descriptors.
%
With recent advances in geometric machine learning~\citep{Wang2019DynamicGC,Qi2017PointNetDL},
researchers extend traditional framework by replacing hand-crafted descriptor with learnable feature descriptors~\citep{Litany2017DeepFM,Fan2019BoostingLS,Zhou2016LearningDC}.
\citet{halimi2019unsupervised,Eisenberger2021NeuroMorphUS} further mitigate the requirements of correspondences annotations 
which builds soft correspondence matrix $\prod \in \mathbb{R}^{m\times n}$ between numerable vertices on the mesh surface.
The lack of explicit surface and numerable vertices in NeRF hinders the use of above methods,
where correspondence affinity matrix~\citep{Eisenberger2021NeuroMorphUS} could not be built.
Pioneer works~\citep{learning-implicit-functions-for-topology-varying-dense-3d-shape-correspondence,deng2021deformed,zheng2021deep} 
propose to build correspondences over implicit representations.
However,
they still rely on ground-truth reconstruction annotations to train the deformation field.
Collecting such annotations for NeRF-based representations is infeasible,
where there are infinite points with non-zero densities.

\noindent
\textbf{Neural Implicit Representations for 3D Geometry.}
The success of deep learning over 2D domain has spurred a growing interest in the 3D domain.
However,
traditional explicit representation such as mesh and voxel are hard to fit in deep learning optimization framework due to the varying topology or limited resolution.
As a parallel class of shape representation,
recent advances in implicit functions~\citep{mescheder2019occupancy,park_deepsdf_2019,Chen2019LearningIF}
have demonstrated their excellence when representing complicated geometry.
By representing shapes 
as a continuous field,
implicit representation encodes the geometry properties of a 3D point $x$ using a neural network $f(x)$.
Conventional implicit representations were limited by the need of 3D ground-truth.
NeRF~\citep{mildenhall2020nerf} stands out as a successful variant to support direct learning of 3D scene from multi-view images.
~\citet{pumarola_d-nerf_2020,park_deformable_2020,li2020neural,park2021hypernerf,guo2021adnerf} further improve NeRF to model non-rigid and time-varying scenes by equipping static NeRF MLP with an extra deformation field.
~\citet{zheng2022imavatar,hong2021headnerf,2021narf,Wang_2021_CVPR,Gafni_2021_CVPR} augment NeRF MLP with a template shape using 3D basic models,
including 3DMM~\citep{hong2021headnerf,Gafni_2021_CVPR}, FLAME~\citep{zheng2022imavatar} and SMPL~\citep{2021narf} to enable more explicit control.
However, they are still limited to overfitting setting and the learned models fail to generalize to novel scenes.
Please note that implicit shape representation and neural rendering are still developing rapidly and we refer readers to the survey~\citep{Tewari2020NeuralSTAR} for more details.

Though great advances have been achieved, 
building dense correspondence across shapes represented by implicit functions are intrinsically challenging since 
ground truth correspondence are impossible to acquire. 
Recent attempts to build correspondence over implicit representations~\citep{zheng2021deep,deng2021deformed} 
tried to bypass this requirement by defining $F$ as signed distance function (SDF) values of the deformed points and $d$ as the marginal $L_1$ loss 
as in~\citep{park_deepsdf_2019}.
\citet{learning-implicit-functions-for-topology-varying-dense-3d-shape-correspondence} followed similar principles as functional maps and adopted occupancy loss as supervision,
while the basis functions are learned from data.
Though dense correspondence over implicit functions could be derived, 
these methods are unable to establish consistent bijective correspondence and still require 3D supervision during training.
Moreover,
these methods are all constrained on synthetic dataset~\citep{shapenet2015}, 
which limit the applications on real scenes.

Our method is  different from them in three ways.
First, our method builds on NeRF that has been shown more effective in representing realistic scenes.
Second, our method is fully free of 3D annotations like sparse correspondence labeling or 3D models.
This uniqueness of our method facilitates more downstream applications where only 2D images are available.
Lastly, our method builds bijective correspondences between two NeRFs,
offering more flexibility and scalability to deform between two NeRFs.

\noindent
\textbf{Generative Models and 3D-aware Image Synthesis.}
Deep generative models, especially GANs~\citep{Goodfellow2014GenerativeAN,Karras2019ASG,Brock2019LargeSG}, have shown promising results in generating photorealistic images.
To further extend GANs to synthesize images in a 3D-consistent manner,
many recent approaches investigated how to incorporate 3D inductive bias into generative training.
Motivated by the success of NeRF~\citep{mildenhall2020nerf},
pioneer work~\citep{Chan2021piGANPI,Schwarz2020NEURIPS}
resorted to the continuous power of radiance fields as the incorporated 3D inductive bias in GANs,
which have paved the way for this field.
Impressive results have been achieved on both 3D-aware image synthesis and multi-view consistency.
More 3D-aware GANs~\citep{gu2022stylenerf,orel2021stylesdf,Chan2021,GIRAFFE,zhou2021CIPS3D} are proposed to support faster rendering~\citep{GIRAFFE}, 
better shape modeling~\citep{orel2021stylesdf,Chan2021}, as well 3D style transfer~\citep{zhou2021CIPS3D}.
%
Without loss of generality,
here we employ the basic $\pi$-GAN~\citep{Chan2021piGANPI} architecture 
as both a robust correspondence similarity metric and an infinite source of 3D NeRFs. 
Beyond the study of improving the synthesis quality,
few work probes how to apply the representations learned by GANs for downstream tasks. Some studies
\citep{bau2020units,Shen2020InterFaceGANIT} interpret the semantics encoded by GANs and apply them for image editing. 
Other works
\citep{StyleGAN3D,Tritrong2021RepurposeGANs,Zhang2021DatasetGANEL} leverage the rich semantics in GAN's features for fine-grained annotation synthesis, few-shot segmentation as well as multi-view data generation, respectively.
Concurrently \citet{pan_2d_2020,Eslami2018NeuralSR,gansteerability,StyleGAN3D}
show that GAN trained on 2D images can learn implicit notion of 3D environment.
But it remains much less explored whether the learned GAN representations are transferable to more challenging 3D tasks, 
like dense correspondence estimation.



\section{Methodology}
\label{sec:methodology}
In this paper,
we present a new attempt for building dense correspondence between NeRF representations across objects belonging to the same category.
Obtaining ground-truth correspondence annotations is infeasible due to the implicit nature of NeRF.
Our key insight is to retrofit a generator of a pre-trained NeRF-based $\pi$-GAN, denoted as $G$, into triple roles:
1) the latent codes in $G$ serve as holistic global structure indicators that improve the generality of models;
2) the features of $G$ serve as geometric-aware local descriptors that enable a feature-based learning objective;
3) and the manifold of $G$ serves as a source of infinite training and evaluation samples 
over a single category.

In the following sections,
we first introduce the details of NeRF-based $\pi$-GAN in Sec.~\ref{sec:method:background} as the background knowledge for subsequent sections.
Next,
we explain the problem formulation and our framework in Sec.~\ref{sec:method:ddf}, learning objective in Sec.~\ref{sec:method:training_obj}, and training strategy in Sec.~\ref{sec:method:training_strategy}.

\subsection{Background on NeRF-based GANs}
\label{sec:method:background}
Inspired by the success of NeRF as an efficient 3D representation,
NeRF-based GANs employ NeRF as their internal representation for 3D-aware image synthesis.
We adopt $\pi$-GAN~\citep{Chan2021piGANPI} in this paper.
Specifically,
the generator of the $\pi$-GAN contains a mapping network $\gM$ and a multi-layer perceptron (MLP) network.
Starting from a latent code $\code\sim p_{Z}$ that follows the Gaussian prior distribution,
the mapping network first maps $\code$ to a set of modulation signals $\gM(z)=\{\bm{\beta}, \bm{\gamma}\}$, where $\bm{\beta}=\{\beta_i\}, \bm{\gamma}=\{\gamma_i\}$.
In $\pi$-GAN, a NeRF is obtained by the MLP network,
which estimates the view-dependent density $\sigma \in \mathbb{R}^+$ and the color vector $\vc\in\mathbb{R}^3$ for each 3D point, taking its coordinate $\point\in\mathbb{R}^3$ and a viewing direction $\vd\in\mathbb{S}^2$ as input.
To associate a latent code to its corresponding NeRF,
the modulation signals will be injected into the MLP network,
serving as FiLM conditions \citep{perez2018film,dumoulin2018feature-wise,sitzmann2020siren} to modulate its features at different layers as $\vf_{i+1}=\sin (\gamma_{i}\cdot(\mW_{i}\vf_i+\vb_i)+\beta_i)$.

\begin{figure*}[t] 
\centering
\includegraphics[width=\linewidth]{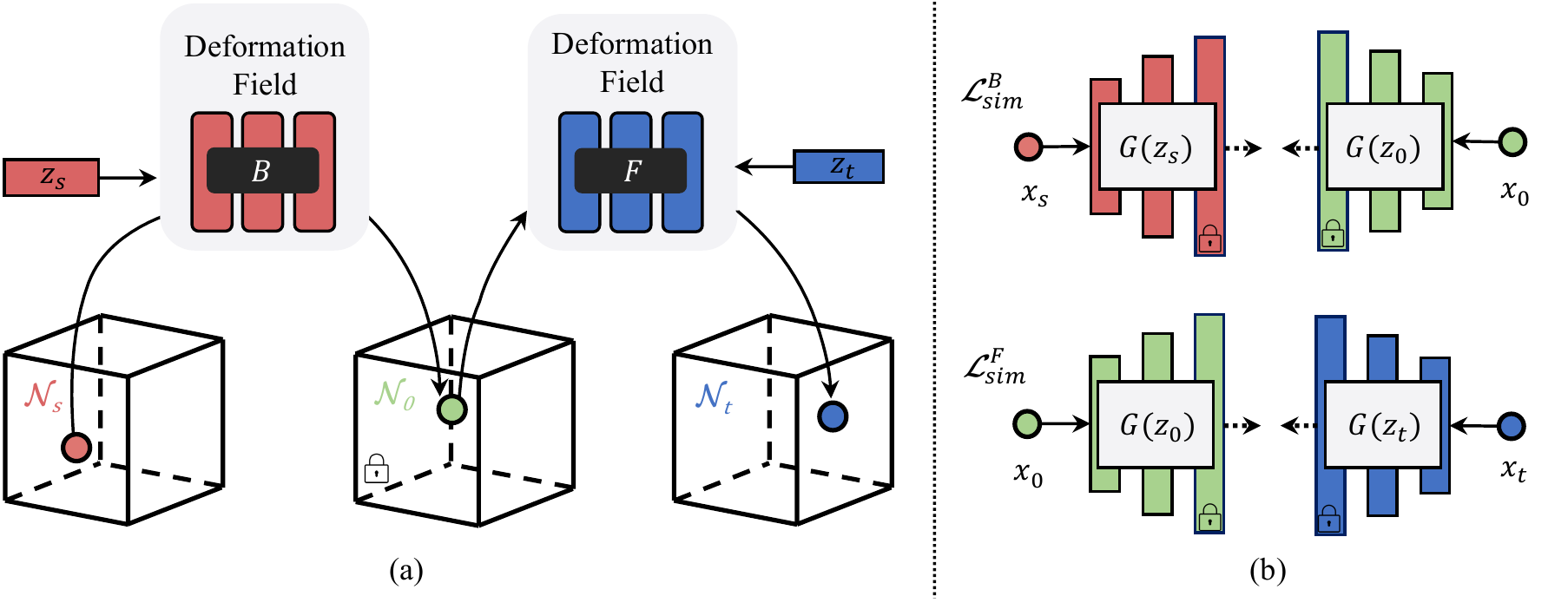}
\vspace{-1cm}
\caption{\textbf{Overview of the proposed Dual Deformation Field (\ECCVMETHODNAME{}).} 
(a) \ECCVMETHODNAME{}~consists of two coordinate-based deformation fields,
namely the backward $B$ and the forward $F$.
To get the correspondence point given a point $\spoint$ sampled from the source NeRF $\nerf_s$,
the $B$ model conditions on the $\code_s$ and learns to deform the input point $\spoint$ to the correspondence point in the template NeRF $\nerf_0$.
Similarly,
the $F$ model conditions on the target latent code $\code_t$ 
and learns to deform points from the template NeRF $\point_0$ to the target NeRF $\nerf_t$.
(b): Feature similarity losses $\mathcal{L}^B_{sim}$ and $\mathcal{L}^F_{sim}$ between features extracted from the generator of the pre-trained $\pi$-GAN, $G$, is adopted as the main loss. 
Please refer to Fig.~\ref{fig:other_loss_detail} for the details of the other two supervisions imposed in the training.
}
\label{fig:overview}
\end{figure*}


Image synthesis in $\pi$-GAN is achieved by sampling a latent code 
and subsequently rendering an image from the corresponding NeRF.
Following the volume rendering of NeRF \citep{mildenhall2020nerf},
each pixel color $C$ of the image is obtained via sampling a set of points along the ray $r(t) = \vo+t\vd$ 
and accumulating their color vectors weighted by their transmittance:
\begin{align}
    \hat{C}(\vr) = \sum_{i=1}^{N} T(t_i) (1-\exp(-\sigma_{i}\delta_{i}))\vc_{i},\label{eq:volume_render}
\end{align}
where $T(t) = \exp\left(-\sum_{j-1}^{i=1}\sigma_{j}\delta_{j}\right)$ and $\delta_{i}=t_{i+1}-t_{i}$ is the distance between adjacent samples.
Using a set of unposed 2D images,
$\pi$-GAN is trained progressively with the non-saturating GAN loss and the R1 regularization \citep{Mescheder2018WhichTM}.

\subsection{The Proposed Framework} 
\label{sec:method:ddf}

\noindent\textbf{Problem Formulation.}
Given any pair of NeRFs $\nerf_{s}:\mathbb{R}^3 \mapsto \mathbb{R}^4$ and $\nerf_{t} :\mathbb{R}^3 \mapsto \mathbb{R}^4$, 
our goal is to estimate a 3D deformation residual $H_D: \mathbb{R}^3 \mapsto \mathbb{R}^3 $ that deforms NeRF 
$\nerf_{s}$ towards NeRF $\nerf_{t}$ via:
\begin{align}\label{eq:deformation_formulation}
    \nerf_{s} \xrightarrow{} \nerf_{t}: \point_t = (\point_s+H_D(\point_s)), \forall \point_{s} \in \nerf_{s}.
\end{align}
The deformation field $H_D$ represents the residual 3D deformation $D(\point_s) = \Delta{\point_s}$ in the 3D space of the source NeRF $\nerf_{s}$.
It is an injective mapping that maps each 3D point $\point_{s}$ in the source NeRF, $\nerf_{s}$, to its corresponding position in the target NeRF, $\nerf_{t}$.

\noindent\textbf{Challenges.}
The problem formulation shown above follows existing attempts~\citep{zheng2021deep,deng2021deformed} that 
model the dense 3D correspondences between an SDF shape and a shared template via a single deformation field.
However, this design does not suit NeRF for the following reasons.
First,
their parameterization is designed to facilitate shape reconstruction, rather than establishing correspondences between two existing shapes.
Second,
deforming all the points on a shape to a shared template could only guarantee an injective mapping instead of a bijective mapping,
where a random point over the template could not find its correspondence on a target shape.
Third,
this design limits information (\eg,~textures) propagation between NeRFs.
Given a ray that intersects with a shape,
unlike SDF representation where the shape surface is modeled by a single point on the zero-level iso-surface,
the volume-based representation (\eg,~NeRF) represents the shape boundary by innumerable points~\citep{zhang2020nerf++}.
Therefore,
after the source NeRF, $\nerf_{s}$, deforms densely sampled near-surface points with texture information to the template,
it is computationally intractable for the target NeRF, $\nerf_{t}$, to find the precise corresponding texture for points along a ray.

\noindent\textbf{Dual Deformation Field.}
We propose to fix the above-mentioned issues by lifting the injective mapping to a bijective mapping function.
A straightforward solution here is to leverage a single
conditional mapping function 
${D}: \mathbb{R}^3 \times \mathbb{R}^{\code_{t}} \times \mathbb{R}^{\code_{s}} \mapsto \mathbb{R}^3$, 
which estimates the offset for each point $\point$ of the source NeRF $\nerf_{s}$,
taking its coordinate and the latent codes $\code_{t}$ and $\code_{s}$ of target and source NeRFs as input.
However, 
since the source and target NeRFs vary in each iteration,
such a solution requires a large model capacity and fails to converge in practice.
A similar observation has also been proposed in previous work that models dynamic NeRF~\citep{pumarola_d-nerf_2020}.
 
To alleviate the computational complexity without sacrificing the bijective property, as illustrated in Fig.~\ref{fig:overview}, we sample a fixed NeRF with a latent code $\code_0$ from $G$ as the intermediate template $\nerf_0$,
and reformulate the deformation field ${D}$ as the composition of two separate conditional neural deformation fields,
namely, a backward deformation field $B$ that estimates the deformation from a source NeRF, $\nerf_{s}$, to the template $\nerf_0$,
and a forward deformation field $F$ that estimates the deformation from the template $\nerf_0$ to a target NeRF, $\nerf_{t}$.

\begin{figure}[t]
  \centering
  \includegraphics[width=\linewidth, ]{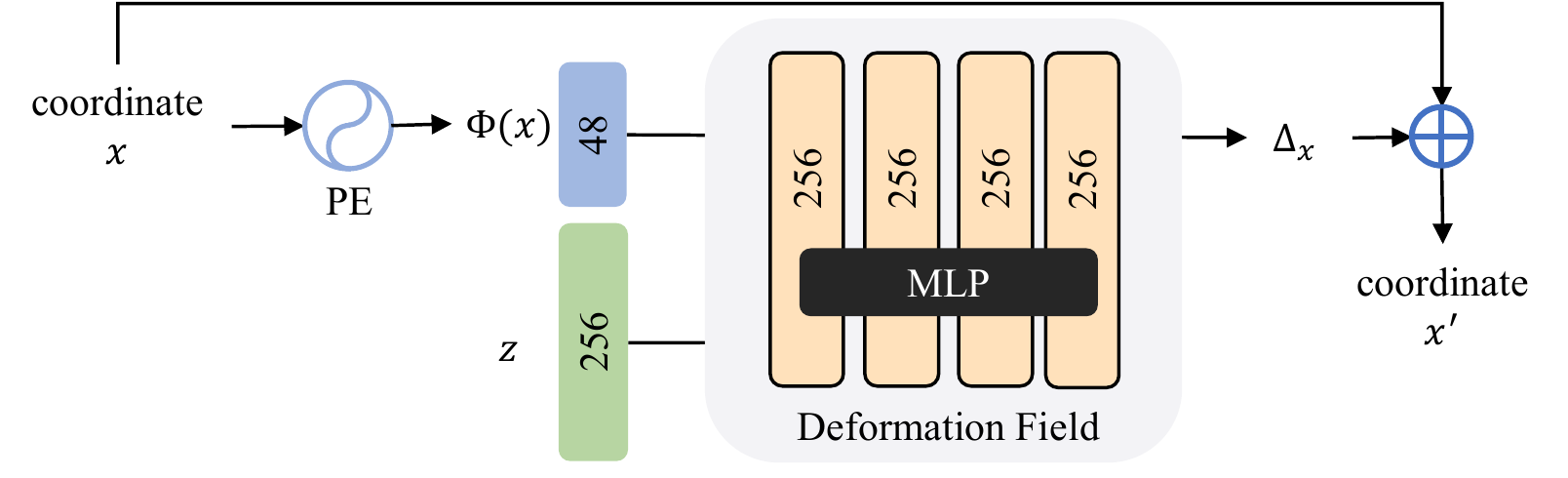}
  \caption{
  \textbf{A diagram of the deformation field model architecture.} 
  Both the forward deformation field $F$ and the 
  backward deformation field ${B}$ 
  are implemented as MLPs consisting of four fully-connected 
  layers with residual connections \citep{mildenhall2020nerf,yu2021pixelnerf}.
  Both $F$ and $B$ take a latent code $\code$ of $256$ dimensions and a coordinate as input, 
  where the latter is embedded into a $48$-dimensional vector via positional encoding~\citep{tancik_fourier_2020,mildenhall2020nerf} 
  } \label{fig:deform_model}
\end{figure}

By decomposing the deformation field between two arbitrary NeRFs into two fields ${B}$ and ${F}$ bridged by a fixed template NeRF, the overall learning complexity is significantly reduced. 
In this way we have
\begin{align}\label{eq:deformation_field}
    \point_0 &= B(\point_s, \code_s), \qquad
    B(\point_s, \code_s) \coloneqq \spoint + H_{B}(\phi(\point_s), \code_{s}),
    \\
    \point_{t} &= F(\point_0, \code_{t}),  \qquad
    F(\point_0, \code_t) \coloneqq \point_0 + H_{F}(\phi(\point_0), \code_{t}), 
\end{align}
where $\spoint \in \nerf_{s}$, $\tpoint \in \nerf_{t}$, and $\point_{0} \in \nerf_{0}$. 
And $\phi(\point)$ is the positional encoding~\citep{mildenhall2020nerf} of a given point.
$H_{B}$ and $H_{F}$ are residual functions each implemented as an MLP consisting of four fully-connected layers, 
as depicted in Fig.~\ref{fig:deform_model}.
The correspondence point of $\spoint \in \nerf_{s}$ in a target NeRF $\nerf_t$ can be retrieved by the composite mapping 
$F(B(\spoint, \code_{s}), \code_{t})$, as depicted in Fig.~\ref{fig:overview}.
%
The latent codes 
$\code_{s}$ and $\code_{t}$ 
serve as the holistic global structure indicators to guide the deformation.
Implementation wise, 
the template NeRF $\nerf_0$ is chosen as 
$\left(
\overline{\bm{\gamma}},\overline{\bm{\beta}}\right)$
which can be intuitively seen as the average shape of the trained dataset.

\subsection{Training Objective}
\label{sec:method:training_obj}
Our overall training objective contains a feature similarity loss for estimated correspondences and three additional regularizations for the deformation fields $F$ and $B$,
namely a cycle-consistency regularization, a second-order feature similarity loss, and a deformation smoothness regularization.

\begin{figure*}[t] 
\centering
\includegraphics[width=\textwidth]{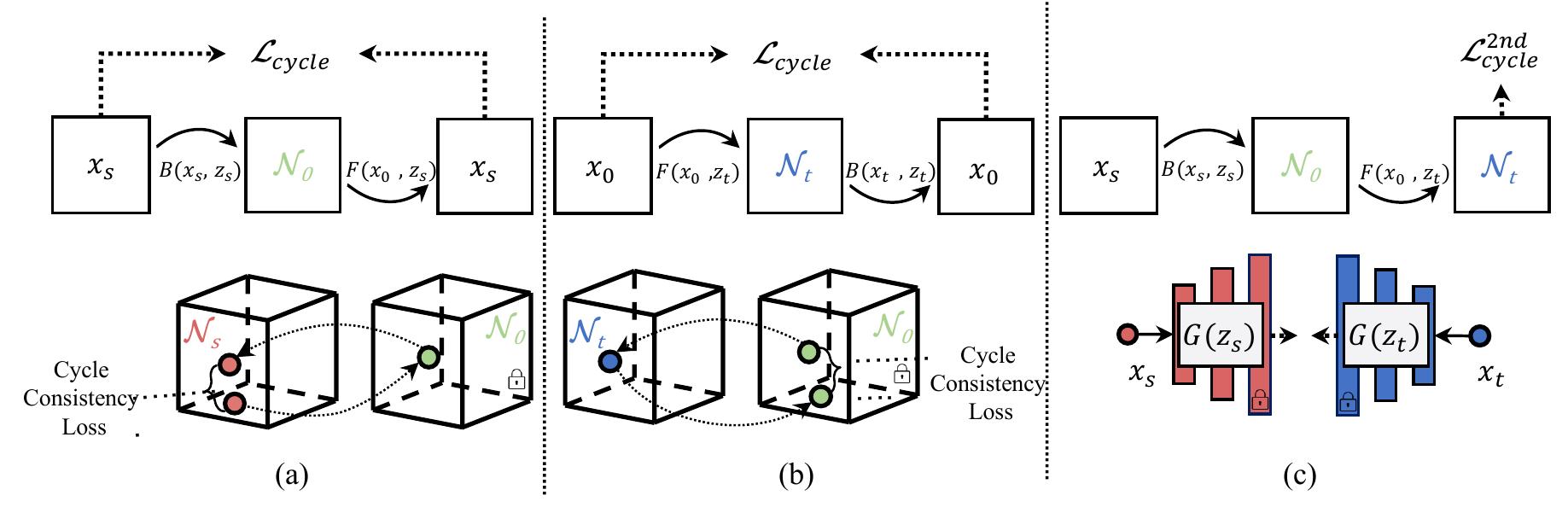}
\centering
\vspace{-1cm}
\caption{
\textbf{Illustration of loss functions used in \ECCVMETHODNAME{}.}
(a) Backward cycle-consistency loss: $F(B(\spoint, \code_s), \code_s) \approx \spoint$,
(b) forward cycle-consistency loss: $B(F(\tpoint, \code_t), \code_t) \approx \tpoint$
and (c) second-order feature similarity loss: $G(\spoint, \code_s) \approx G(F(B(\spoint, \code_s),\code_t),\code_t)$.
}
\label{fig:other_loss_detail}
\end{figure*}

\noindent
\textbf{Generator Feature Similarity Loss.} 
\label{sec:method:feat_sim_loss}
Given a collection of $n$ source NeRFs $\{\nerf^{(i)}_s\}_{i=1}^{n}$ that are sampled from 
$G$ with corresponding latent codes $\{\code^{(i)}_s\}_{i=1}^n$,
each of these NeRFs will serve as a source NeRF for ${B}$ to compute its deformation to the template.
For each pair of estimated corresponding points $(\point_{s},\point_{0})$ where $\point_{s}$ belongs to one of these source NeRFs and $\point_{0}$ belongs to the template,
we take a point feature extracted from NeRF generator $G$ as the local geometric descriptor.
When the $\point_{s}$ and $\point_{0}$ are homologous and share similar semantic meanings, 
the feature similarity loss should be small,
and a smaller feature similarity loss in training indicates that the deformation field produces reasonable correspondences.
Therefore, 
for each pair of sampled points, we compare the cosine similarity between two descriptors as their correspondence relevance score and update the network accordingly.
%
%
Consequently,
the feature similarity loss for $B$ can be written as:
\begin{equation}
\label{eq:L_similarity_B}
\begin{split}
    \mathcal{L}_{sim}^{B} = 
    &
    \frac{1}{n}
    \sum_{i=1}^{n}
    \Bigg[
    \frac{1}{\lvert{\mathcal{P}^{(i)}_{s}\rvert}}
    \sum_{\point_s \in \mathcal{P}^{(i)}_s}  \\
    &
    \textcolor{black}{w_{\point_{s}}} * 
    \frac{1}{2} 
    \norm{
    (G(\spoint ,\code_s^{(i)}) - G(B(\spoint, \code_s^{(i)}),\code_0))
    }_2^{2}
    \Bigg],
\end{split}
\end{equation}
where the loss of each point $\point_{s}^{(i)}$ is weighted by 
$w_{\point_{s}}=T(t_{\point_{s}})$ defined in Eq.~\ref{eq:volume_render},
so that $B$ is encouraged to focus more on points with large densities,
as they are close to the object surface with rich semantics.
It is worth noting that to reduce computational redundancy and complexity,
we will sample only a subset $\mathcal{P}^{(i)}_s$ of points 
from each NeRF $\nerf^{(i)}_s$ by the sampling strategy introduced in the next section.
Each of these NeRFs will also serve as a target NeRF 
for $F$ to compute the deformation of the template to it.
The feature similarity loss for $F$ is thus:
\begin{equation}
\label{eq:L_similarity_F}
\begin{split}
    \mathcal{L}_{sim}^{F} = 
    &
    \frac{1}{m}
    \sum_{j=1}^{m}
    \Bigg[
    \frac{1}{\lvert{\mathcal{P}^{(j)}_0\rvert}} 
    \sum_{\pointtemplate\in\mathcal{P}^{(j)}_0} \\
    &
    \textcolor{black}{w_{\point_{0}}} * 
    \frac{1}{2} 
    \norm{G(\point_{0},\code_0) - G(F(\point_{0},\code_t^{(j)}),\code_t^{(j)})}_2^{2}
    \Bigg],
\end{split}
\end{equation}
where $\pointtemplate$ stands for a point on the template $\nerf_0$ and we sample a subset $\mathcal{P}^{(j)}_0$ from $\nerf_0$ for each different target NeRF $\nerf^{(j)}_t$.
For all the feature similarity supervision,
we adopt features of $G$ at multiple layers and concatenate them to better reflect the semantics of a point.
We justify our choice in Sec.~\ref{sec:exp:abla:feature_analysis}.

\noindent
\textbf{Cycle-Consistency Regularization.}
Since the conditional deformation fields, $F$ and $B$, are supposed to restore the original deformation field $D$,
when the same NeRF $\nerf_i$ is used as both the source and target NeRF,
they should satisfy $D(\point , \code_{t}, \code_{t}) = \point $ for all valid points $\point $.
As depicted in Fig.~\ref{fig:other_loss_detail}(a,b), we further apply a cycle-consistency regularization for $B$ and $F$:
\begin{equation}
\label{eq:cycle_consistency}
    \begin{split}
   & \mathcal{L}_{cycle} = \\
   & \frac{1}{n}
    \sum_{i=1}^{n}
    \Bigg[
    \frac{1}{\lvert{\mathcal{P}^{(i)}_s\rvert}}
    \sum_{\point_s \in\mathcal{P}^{(i)}_s}
    \norm{F(B(\spoint , \code_{s}^{(i)}), \code_{s}^{(i)}) - \spoint }_{2}^{2} \Bigg] + \\
   & \frac{1}{m}
    \sum_{j=1}^{m}
    \Bigg[
    \frac{1}{\lvert{\mathcal{P}^{(j)}_0\rvert}}
    \sum_{\pointtemplate\in\mathcal{P}^{(j)}_0}
    \norm{B(F(\pointtemplate, \code_t^{(j)}), \code_t^{(j)}) - \pointtemplate}_{2}^{2}
    \Bigg].
    \end{split}
\end{equation}

\noindent
\textbf{Second-Order Feature Similarity Loss.}
Apart from the aforementioned point-wise cycle-consistency loss that regularizes the deformation coherency of learned mapping,
we also combine it with Eq.~\ref{eq:L_similarity_B},~\ref{eq:L_similarity_F} and impose a feature-based cross-instance cycle-consistency loss. 
Specifically,
for a given point $\point_s$ in a source NeRF $\nerf_{s}^{(i)}$ paired with latent code $\code_s^{(i)}$,
beyond imposing the similarity regularization only over the template NeRF $\nerf_0$,
we further deform its intermediate point ${\point_0} = B(\point_{s}, \code_{s})$ to a randomly sampled paired target NeRF $\nerf_t^{(i)}$ and calculate their feature similarity:
\begin{equation}
\label{eq:feat_cycle}
\begin{split}
    &
    \mathcal{L}_{cycle}^{2nd} = 
    \frac{1}{n}
    \sum_{i=1}^{n}
    \Bigg[
    \frac{1}{\lvert{\mathcal{P}^{(i)}_s \rvert}} 
    \sum_{\point_s\in\mathcal{P}^{(i)}_s} \\
    &
    \textcolor{black}{w_{\point_s}} * 
    \frac{1}{2} 
    \norm{G(F(B(\point_{s}, \code_{s}^{(i)}), \code_t^{(i)}), \code_t^{(i)}), 
        G(\point_s , \code_{s}^{(i)})}_2^{2}
    \Bigg].
\end{split}
\end{equation}
\noindent
We find this auxiliary regularization improves cross-instance deformation consistency.

\noindent
\textbf{Deformation Smoothness Regularization.} \label{smooth_loss}
To encourage the smoothness of deformation and reduce spatial distortion,
a deformation smoothness regularization is also included.
Here we penalize the norm of the Jacobian matrix 
$\mathbb{J}_{D} = \nabla {D}$ 
of the deformation fields 
~\citep{park2021nerfies}
to ensure the learned deformations are physically smooth:
\begin{equation}
\label{eq:smoothness}
\begin{split}
    &
    \mathcal{L}_{smooth} = \\  
    &
    \frac{1}{n}
    \sum_{i=1}^{n}
    \Bigg[
    \frac{1}{\lvert{\mathcal{P}^{(i)}_s\rvert}}
    \sum_{\point_s  \in\mathcal{P}^{(i)}_s}
    \max(
    \norm{
   \nabla B(\point_s, \code_s^{(i)}) 
    }_{2}^{2}-\epsilon, 0) \Bigg] +\\
    &
    \frac{1}{m}
    \sum_{j=1}^{m}
    \Bigg[
    \frac{1}{\lvert{\mathcal{P}^{(j)}_0\rvert}}
    \sum_{\point_0\in\mathcal{P}^{(j)}_0}
    \max(
    \norm{
    \nabla F(\point_0, \code_t^{(j)})
    }_{2}^{2} - \epsilon, 0
    \Bigg],
\end{split}
\end{equation}
\noindent
where $\epsilon$ is the slack parameter for the smoothness regularization.
The final objective is thus $\mathcal{L}_{total} = \mathcal{L}_{sim}^F + \mathcal{L}_{sim}^B + \lambda_{cycle} \mathcal{L}_{cycle} 
+ \lambda_{cycle}^{2nd} \mathcal{L}_{cycle}^{2nd}
+ \lambda_{smooth} \mathcal{L}_{smooth}$ 
where $\lambda_{cycle}$, $\lambda_{cycle}^{2nd}$, and $\lambda_{smooth}$ are balancing coefficients, which are respectively set to $1$, $0.1$ and $10^{-4}$ in practice. 

\subsection{Training Strategy}
\label{sec:method:training_strategy}
While the pre-trained $\pi$-GAN $G$ serves as a source of infinite object NeRFs,
in each iteration of the training process we will sample a batch of source NeRFs $\{\nerf_s^{(i)}\}_{i=1}^{n}$ with corresponding latent codes $\{\code_s^{(i)}\}_{i=1}^n$,
and a batch of target NeRFs $\{\nerf_t^{(j)}\}_{j=1}^{m}$ with the corresponding latent codes $\{\code_t^{(j)}\}_{j=1}^m$.
To further sample a point set 
for each sampled NeRF
$\nerf_{*}^{(i)}$,
for each pixel within the resolution ${H\times W}$
we shoot a ray $r(v) = \vo+v\vd$ where $\vd$ identifies the direction from the camera to the pixel.
Subsequently, 
for each ray we follow \citet{mildenhall2020nerf}  
and conduct a hierarchical sampling to obtain a \emph{fine} set of points, 
\ie, points near the object surface.
We denote the union of these point sets sampled from source as $\{\mathcal{P}^{(i)}_s\}_{i=1}^{n}$,
which are used to train the $B$ model.
Since the points sampled to train $F$ models are all from the template NeRF $\nerf_0$,
here we denote the the point sets paired with target NeRF $\nerf_t^{(j)}$ as $\{\mathcal{P}^{(j)}_0\}_{j=1}^{m}$ for clarity.

\noindent
\textbf{Curriculum Sampling of NeRFs.} 
In practice,
we find the variation between the sampled NeRF and the template NeRF can significantly affect the training process,
which may even collapse at the beginning stage if it gets a sampled NeRF that differs substantially from the template.
 
To improve training stability and efficiency,
we adopt a curriculum sampling strategy to obtain NeRFs from $G$,
by gradually morphing the template NeRF in the latent space to sample NeRFs with growing complexity.
Specifically,
since in $\pi$-GAN, the semantics of a sampled NeRF is determined by the modulation signals $(\bm\beta,\bm\gamma)$,
we can linearly interpolate between two sets of modulation signals to gradually morph one NeRF into another.
Inspired by this property of $\pi$-GAN,
when we sample a set of $n$ NeRFs $\{\nerf^{(i)}\}_{i=1}^{n}$,
we compute their corresponding modulation signals $\{(\bm\beta^{(i)},\bm\gamma^{(i)})\}_{i=1}^n$ from their latent codes.
Subsequently,
we adjust the learning complexity by blending them with the template NeRF as
\begin{align}\label{curriculum_interp}
\bm\beta^{(i)}(\alpha) & = \bm\beta_0 + \alpha \cdot(\bm\beta^{(i)} - \bm\beta_0) \\
\bm\gamma^{(i)}(\alpha) & = \bm\gamma_0 + \alpha \cdot(\bm\gamma^{(i)} - \bm\gamma_0),
\end{align}
where $(\bm\beta_0, \bm\gamma_0)$ are the modulation signals of the template NeRF and $\alpha$ controls the learning difficulty.
In practice, we start from $\alpha=0$ and linearly increase the value to $0.6$ during training,
which is a reasonable value to balance sampling quality and diversity~\citep{karras2019style}.
In this way,
the model learns to produce identity deformation first and then gradually evolves to model more complicated deformation when trained on more challenging samples.
%


\section{Experiments}
\label{sec:experiments}
\subsection{Experimental Setup}
\begin{table*}[tp]
\caption{Hyper parameters of the sampling and regularization loss weights.}
\centering
    \footnotesize
    \begin{threeparttable}
    \begin{tabular}{lcccccc} 
        \toprule
        Dataset & Ray Steps & Depth Mask & Sampling Ratio & Batch Size & $\lambda_{cycle}$ & $\lambda_{smooth}$\\ 
        \midrule 
        CelebA~\citep{liu2015faceattributes}     & 24 & 1.08 & 0.2 & 131,072 & 0.1 & 0.1\\
        Carla~\citep{Schwarz2020NEURIPS} & 48 & 1.2 & 0.05 & 65,536 & 0.05 & 0.01 \\
        Cats~\citep{cats}     & 36 & 1.08 & 0.1 & 49,152 & 0.1 & 0.1 \\ 
        \bottomrule
    \end{tabular}
    \end{threeparttable}
    \vspace{2pt}
    \label{tab:exp:sampling_detail}
\end{table*}
As discussed in~\citep{deng2021deformed,zheng2021deep}, 
there is no ground-truth dense correspondence dataset available for structure with variations.
Therefore, we adopt three proxy tasks as surrogate metrics to evaluate the learned correspondences of \ECCVMETHODNAME{}.
In Sec.~\ref{exp:subsec:texture_transfer}, 
we first qualitatively 
demonstrate the dense correspondences learned by \ECCVMETHODNAME{}~through 
texture transfer. 
Quantitative results are shown in two alternative tasks, 
namely fine-grained segmentation transfer and keypoints transfer, in Sec.~\ref{exp:subsec:quant_result_segment} and Sec.~\ref{exp:subsec:quant_result_keypoint}, respectively.
All imagery results shown are rendered at $256^2$ resolution. 

\noindent
\textbf{Training.}
In all the experiments,
we set the learning rate to $5\times10^{-5}$ 
and decay in every $5,000$ iteration with gamma=$0.5$.
We adopt Adam~\citep{Kingma2015AdamAM} optimizer to train the deformation models.
In each training iteration,
we randomly sample a batch of $10$ source NeRFs $\{\nerf_s^{(i)}\}_{i=1}^{10}$ 
with corresponding latent codes $\{\code_s^{(i)}\}_{i=1}^{10}$,
and a batch of target NeRFs $\{\nerf_t^{(j)}\}_{j=1}^{10}$ with corresponding latent codes $\{\code_t^{(j)}\}_{j=1}^{10}$.
For all experiments, 
we train the \ECCVMETHODNAME{} for $80,000$ iterations,
which takes about $8$ hours on a single Tesla V100 GPU.
The hyperparameter details are listed in Tab.~\ref{tab:exp:sampling_detail}.
 
\noindent
\textbf{Evaluation Dataset.}
We extensively demonstrate our approach on human faces~\citep{liu2015faceattributes} as the main object category,
as human faces are rich in geometric details, 
making them the best choice for demonstrating the accuracy, smoothness, and robustness of learned correspondences.
Moreover, human faces are also rich in downstream tasks, from which we can effectively investigate the potential of learned correspondences.
The qualitative results on cats \citep{cats} and cars \citep{Schwarz2020NEURIPS,DBLP:journals/corr/abs-1711-03938} 
are also included.

\noindent
\textbf{Sampling Details.}
To train the \ECCVMETHODNAME{}~network efficiently, 
we conduct hierarchical sampling to obtain 3D point sets with more specific semantic meanings.
As in~\citep{mildenhall2020nerf}, we first uniformly sample points in 3D space and then sample via importance sampling a more informed fine point set given the density output of the ``coarse'' point set.
These samples are biased toward the more relevant parts of the rendered object.
We list the sampling details in Tab.~\ref{tab:exp:sampling_detail}.
Apart from applying a foreground depth mask to filter out background information to increase sampling efficiency,
we also control the sampling ratio of remaining rays.
By defining a smaller sampling ratio,
we could increase the number of NeRF sampled per batch to increase the diversity of training samples.
We curtail the sampling points to a certain number so to maintain the stability of training.

\noindent
\textbf{NeRF-based GANs.}
We use the officially released $\pi$-GAN pretrained models for dense correspondence learning.
To extract network features, 
we use the features starting from layer 4.
We find the middle layer features have more correlation with the underlying semantics of a given region,
while the last few layers are more sensitive to low-level details such as color variations, 
which could not provide meaningful clues for dense correspondence learning. 
We further justify our choice in Fig.~\ref{sec:exp:abla:feature_analysis}.



\begin{figure*}[tp]
  \centering
  \includegraphics[width=1\linewidth, ]{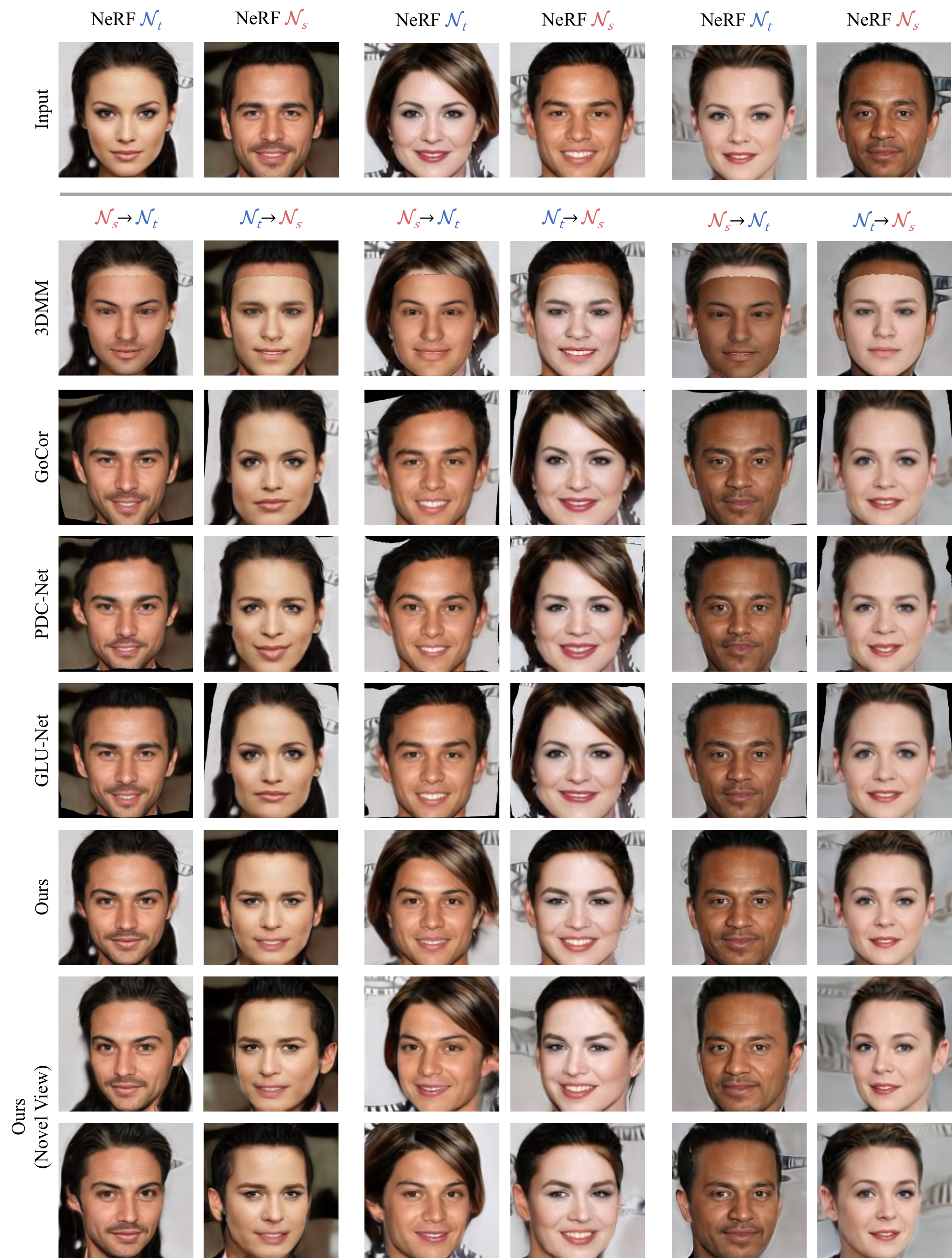}
\caption{\textbf{Texture transfer through the learned deformation field}. 
We randomly sample three NeRF pairs here for qualitative evaluation (shown at the top row as inputs).
For each NeRF sample $\NeRF_{*}$ we transfer the texture from the paired NeRF according to their 3D dense correspondences.
Specifically, for the column labeled with NeRF $\NeRF_{t(s)}$,
we show the texture transfer results from source NeRF $\NeRF_{s(t)} \rightarrow \NeRF_{t(s)}$.
We conduct dual texture transfer on three pairs (depicted in different separated columns) and show the transferred results over three different angles.
The separate line splits the input, the model-based method's output, the learning-based methods' output, and ours.
Though not designed for 2D images,
our method consistently outperforms the baseline method in terms of fidelity and naturalness.
}
\label{fig:texture_transfer}
\end{figure*}
\subsection{Qualitative Results on Texture Transfer}
\label{exp:subsec:texture_transfer}
In this subsection, 
we qualitatively demonstrate the dense correspondences learned by \ECCVMETHODNAME{}~through texture transfer. 
The results here validate that \ECCVMETHODNAME{}~learns accurate underlying structures of NeRFs and their associated correspondences without explicit correspondence supervision provided during training. 
\begin{figure*}[tp]
  \centering
  \includegraphics[width=1\linewidth, ]{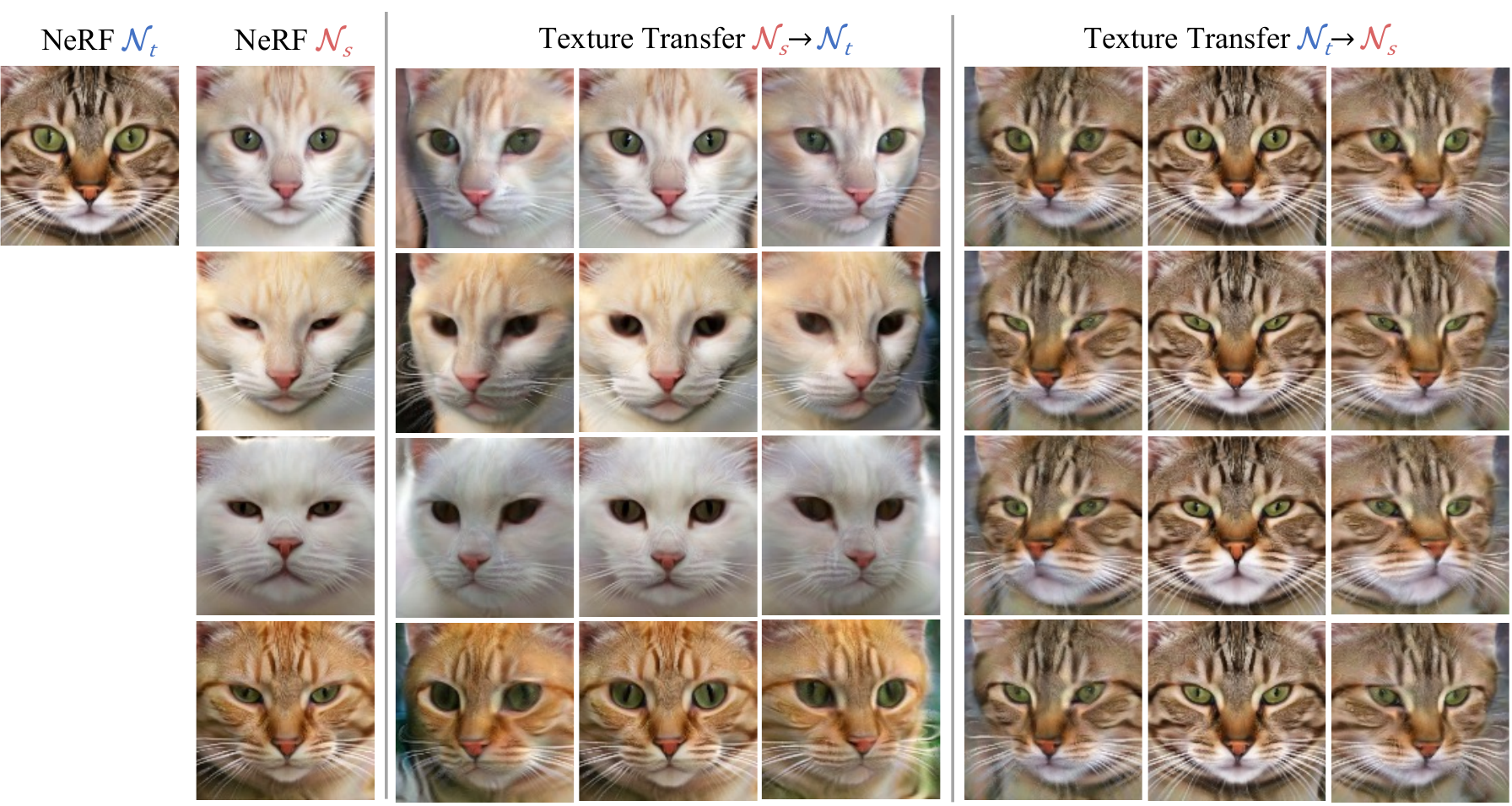}
  \caption{
  \textbf{Visualization of texture transfer on Cats Dataset.}
  The size of the eyes and positions of the nose and the overall shape could serve as hints to observe the difference between different cats.
  } \label{fig:tex_tran_cats}
\end{figure*}

\noindent
\textbf{Texture Transfer via \ECCVMETHODNAME{}.}
We denote $\nerf_s \rightarrow \nerf_t$ as the process of transferring texture from NeRF $\nerf_s$ to NeRF $\nerf_t$ while maintaining the geometry of $\nerf_t$.
To perform the transfer, for each sampling point $\spoint$ to render NeRF $\nerf_s$,
we first deform $\spoint$ to the template correspondence $\point_0$ via $B$
and then deform it to the target NeRF $\nerf_t$ space correspondence point $\tpoint^{\prime}$ via $F$.
We query the geometry of $\nerf_t$ and texture of $\nerf_s$ to conduct volume rendering in the given view direction.
To remove ambiguity, 
we mask out the hair and background class of the source class using segmentation masks and conduct texture transfer on other semantic regions on the human face.

\noindent
\textbf{Results on Human Faces.}
With the above rendering process, we show the cross-instance texture transfer results in Fig.~\ref{fig:texture_transfer}.
We compare our method with two types of baselines,
the model-based 3DMM~\citep{blanz1999morphable} method (row 1)
and the state-of-the-art learning-based 2D correspondence matching methods~\citep{Truong2020GLUNetGU,Truong2020GOCorBG,Truong2021LearningAD} (rows 2-4).
Visually inspected, our method produces semantic plausible dense correspondences with high-fidelity texture transfer results.
We also show our results in multiple views to demonstrate that our method has learned both 3D consistent dense correspondences. 
Note that good texture transfer results could not be achieved without accurate correspondence matching in 3D space. Our approach shows superior texture transfer in comparison to existing model-based and learning-based methods.

\begin{figure*}[tp]
  \centering
  \includegraphics[width=1\linewidth, ]{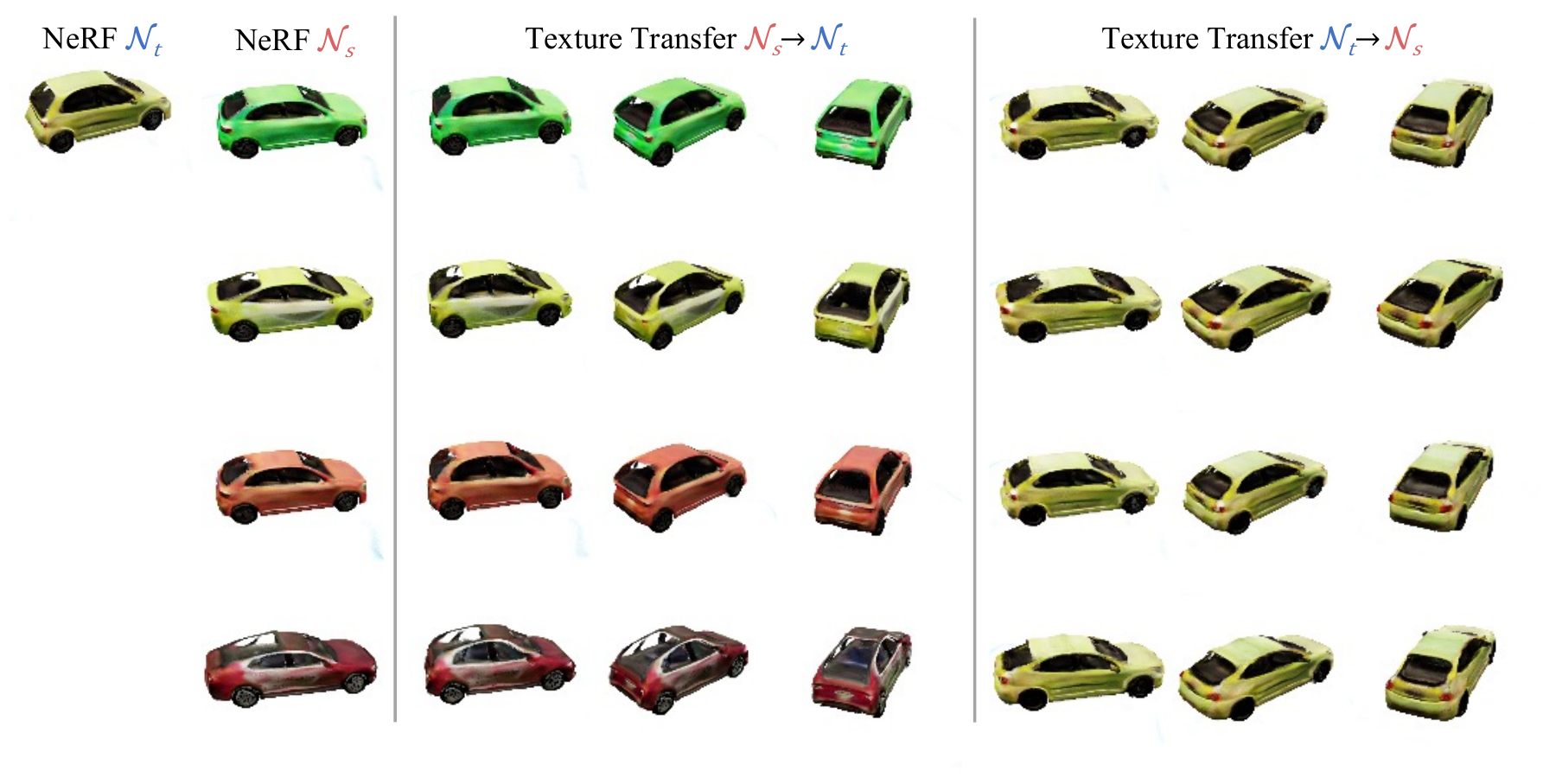}
  \caption{
  \textbf{Visualization of texture transfer on CARLA Dataset.}
  In the category with large structure deviations, 
  \ECCVMETHODNAME{} could still generate sound deformation with high fidelity and accuracy.
  } \label{fig:tex_tran_carla}
\end{figure*}
\begin{figure*}[h!]
  \centering
  \includegraphics[width=1\linewidth, ]{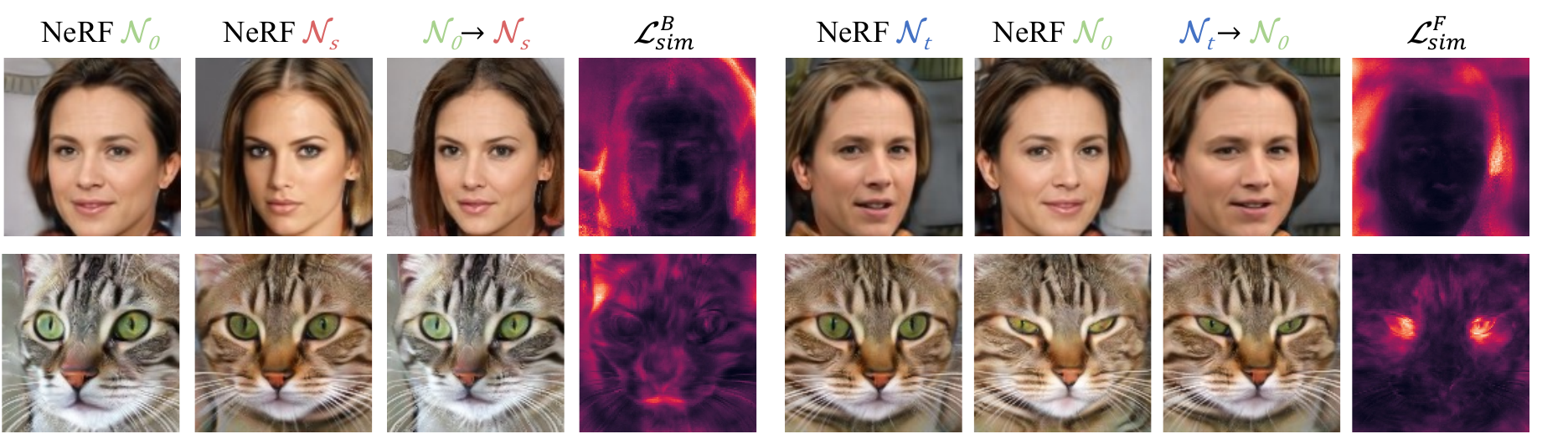}
  \caption{\textbf{Visualization of uncertainty map of learned dense correspondence.}
  The left and the right column shows the pixel-wise uncertainty map corresponding to backward deformation $B$ and forward deformation $F$, 
  respectively.
  The pixel-wise uncertainty is calculated as the integration of the point-wise 
  $\mathcal{L}_{sim}^{*}$ with Eq.~\ref{eq:volume_render}.}
  \label{fig:uncertainty_map}
\end{figure*}

\noindent
\textbf{Results on Other Categories.}
To further illustrate the deformation ability of dual fields in \ECCVMETHODNAME{},
we apply our method on two more pretrained NeRF-based generators, trained respectively on the real-world Cats~\citep{cats} datasets
as well as the synthetic CARLA~\citep{Schwarz2020NEURIPS,DBLP:journals/corr/abs-1711-03938} 
dataset.
We train the corresponding \ECCVMETHODNAME{} models on the new categories with parameters listed in Tab.~\ref{tab:exp:sampling_detail} and conduct texture transfer using the same pipeline.
Given the source NeRF $\nerf_s$ and target NeRF $\nerf_t$, 
we show transfer results from both $\nerf_s \rightarrow \nerf_t$ and $\nerf_t \rightarrow \nerf_s$ to validate the performance of \ECCVMETHODNAME{} on shape categories with larger structure variations.

In Fig.~\ref{fig:tex_tran_cats} 
we show the texture transfer results on the Cats dataset.
Though cat's faces have fewer discriminative features compared to human faces,
through the overall shape and local details such as the size of the cat's eye and mouth, 
we could see that the transferred multi-view results share the same texture with the source NeRF, 
while still matching the geometry of the target NeRF.

In Fig.~\ref{fig:tex_tran_carla} we show the transfer results on synthetic CARLA dataset.
Compared to CelebA and Cat datasets,
Cars have larger structure variations and larger deformations between different NeRFs,
leaving learning accurate deformation on CARLA dataset more challenging. 
Through the qualitative results,
the texture transfer of $\nerf_s \rightarrow \nerf_t$ through \ECCVMETHODNAME{}
produces convincing correspondence across two NeRFs that are largely different.
The shared semantic components are matched to the maximum extent and also preserve the original geometry pattern of NeRF $\nerf_t$.
The texture transfer of the other direction $\nerf_t \rightarrow \nerf_s$ is overall reasonable
but produces mismatches in some regions with large deformations such as the car roof,
whose size varies evidently across different objects represented in NeRFs.

\noindent
\textbf{Uncertainty of Texture Transfer.}\label{sec:exp:ablation:confidence_map}
In Fig.~\ref{fig:uncertainty_map} we showed the uncertainty heat map and the texture transfer results of the 
learned \ECCVMETHODNAME{}. 
After the training of~\ECCVMETHODNAME{},
we conduct correspondence inference and calculate the feature similarity loss of the correspondence points and the original points. 
The feature similarity loss $\mathcal{L}_{sim}^{B}$ and $\gL_{sim}^{F}$ between inferred correspondence points
could be naturally interpreted as the uncertainty of the learned correspondence.
A low feature similarity loss denotes the correctness of deformation and guarantees the visual effects of texture transfer. 
In Fig.~\ref{fig:uncertainty_map},
we separately show the uncertainty maps corresponding to the backward deformation $B$ and the forward deformation $F$.
As can be seen, the semantic regions of the human face have a low uncertainty score, 
except for ambiguous regions like hair.
For Cat face, the overall uncertainty is low except for the regions where 
deformations are large such as cat eyes and mouth.
After~\ECCVMETHODNAME{}~converges,
the heat map could also serve as the confidence score of the dense correspondences between two NeRFs.
%

\subsection{Quantitative Results on Segmentation Label Propagation}
\label{exp:subsec:quant_result_segment}
%
To demonstrate the quantitative performance of the learned dense correspondence, 
following the previous method~\citep{deng2021deformed} we resort to segmentation label propagation as the surrogate metric. 
Intuitively,
a 3D point shall share the same segmentation label with its correspondence point from another object with structure variations if being deformed via an accurate correspondence algorithm.
Thus, segmentation label propagation could serve as a metric to inspect the performance of learned correspondences.
Similar to the texture transfer experiments discussed in Sec.~\ref{exp:subsec:texture_transfer},
here we conduct segmentation label propagation on the fine-grained human faces.

Different from explicit-based representations and SDF-based implicit representation~\citep{park_deepsdf_2019,deng2021deformed,zheng2021deep},
NeRF-based representation is designed for view synthesis and has no clear surface boundary,
leaving it hard to directly evaluate the segmentation accuracy in the 3D space.
Therefore,
we propose to conduct segmentation label propagation in the 3D space and project the propagated labels in the 2D space through volume rendering depicted in Eq.~\ref{eq:volume_render} for evaluation. 
We describe how we conduct segmentation label propagation below.

\begin{figure}[h!]%
    \centering
        \includegraphics[width=1\textwidth]{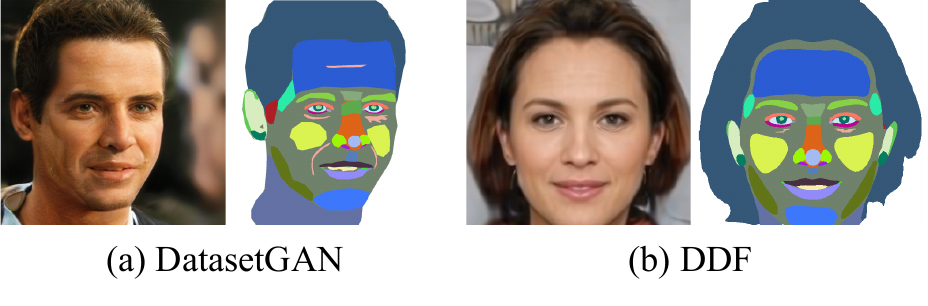} 
        \vspace{-1cm}
        \caption{\textbf{Canonical segmentation annotation for two 1-shot segmenters.}
        (a) DatasetGAN~\citep{Zhang2021DatasetGANEL} and (b) ours.
        For DatasetGAN we choose the first annotated image in their training set,
        and for \ECCVMETHODNAME{}~we simply use the frontal face of the Template for segmentation transfer.
        For ease of comparison,
        the segmentation annotations of the Template are simply acquired through an off-the-shelf pretrained DatasetGAN segmenter,
        which already provides reasonable results. 
        }
         \label{fig:seg-1shot-gt}
\end{figure}

\noindent
\textbf{Segmentation Label Propagation.}
\label{sec:exp:seg_label_prop_details}
For this task,
%
we first render the front view of our template NeRF $\nerf_0$ and provide it with the oracle segmentation map acquired from a pretrained DeepLabV3~\citep{Chen2017RethinkingAC, Zhang2021DatasetGANEL} segmentation model.
We refer to this front view as the oracle image, 
as shown in Fig.~\ref{fig:seg-1shot-gt}(b).
For an unlabeled test image rendered from a NeRF,
for each pixel, 
we cast a ray through this pixel and sample $96$ points along the corresponding ray.
For a point $\point$ along the ray,
we use the network $B$ to query its correspondence point $\vx^\prime$ in the template NeRF.
The projected segmentation label is thus regarded as the segmentation label prediction of $\point$.
To acquire 2D segmentation predictions for evaluation, 
we aggregate the predictions of 3D points 
by rescaling their voting contributions with the transmittance value $T(i)$ defined in Eq.~\ref{eq:volume_render}.
The whole segmentation process costs around $3$ seconds for a single image on a Tesla V100 GPU. 

Since we only use the oracle segmentation map for the oracle image,
we consider our approach as a 1-shot segmentation method.
\begin{figure*}[h]
  \centering
  \includegraphics[width=1.\linewidth,]{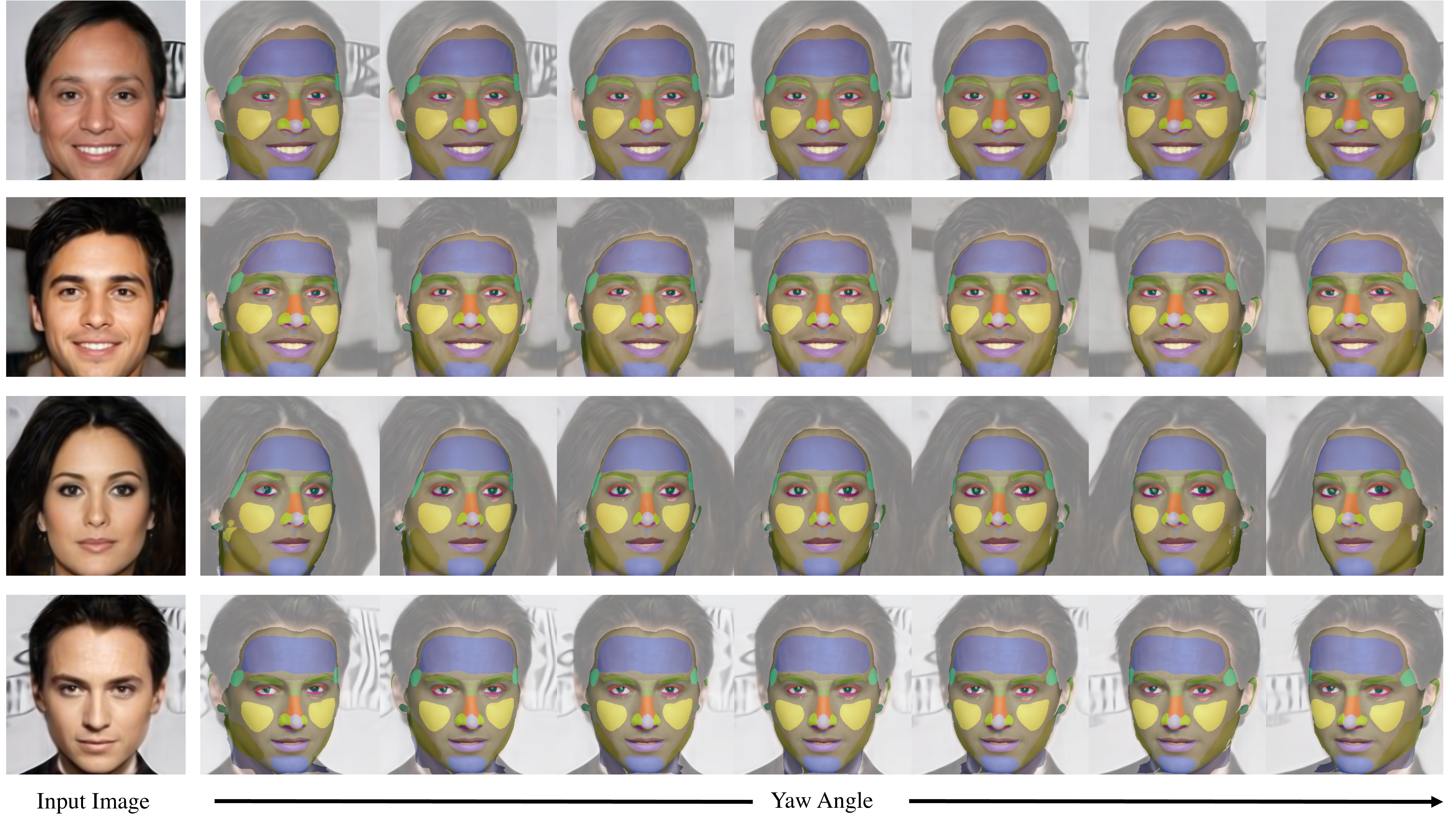}
  \caption{
  \textbf{3D consistent segmentation label transfer with novel poses}.
  Given an annotated projection of Template NeRF shown in Fig.~\ref{fig:seg-1shot-gt}(b), 
  we could derive view-consistent segmentation maps of other NeRF objects through our method.
  Note that for instances missing teeth class (1st row and 2nd row, segment in white color), 
  our method could still derive accurate correspondences though the teeth class does not exist in the segmentation template.
  This demonstrates that our method learns consistent 3D dense correspondence.
  } \label{fig:seg:3d-consistent-synthetic}
\end{figure*}

\begin{figure*}[tp]
  \centering
 \includegraphics[width=1\linewidth, ]{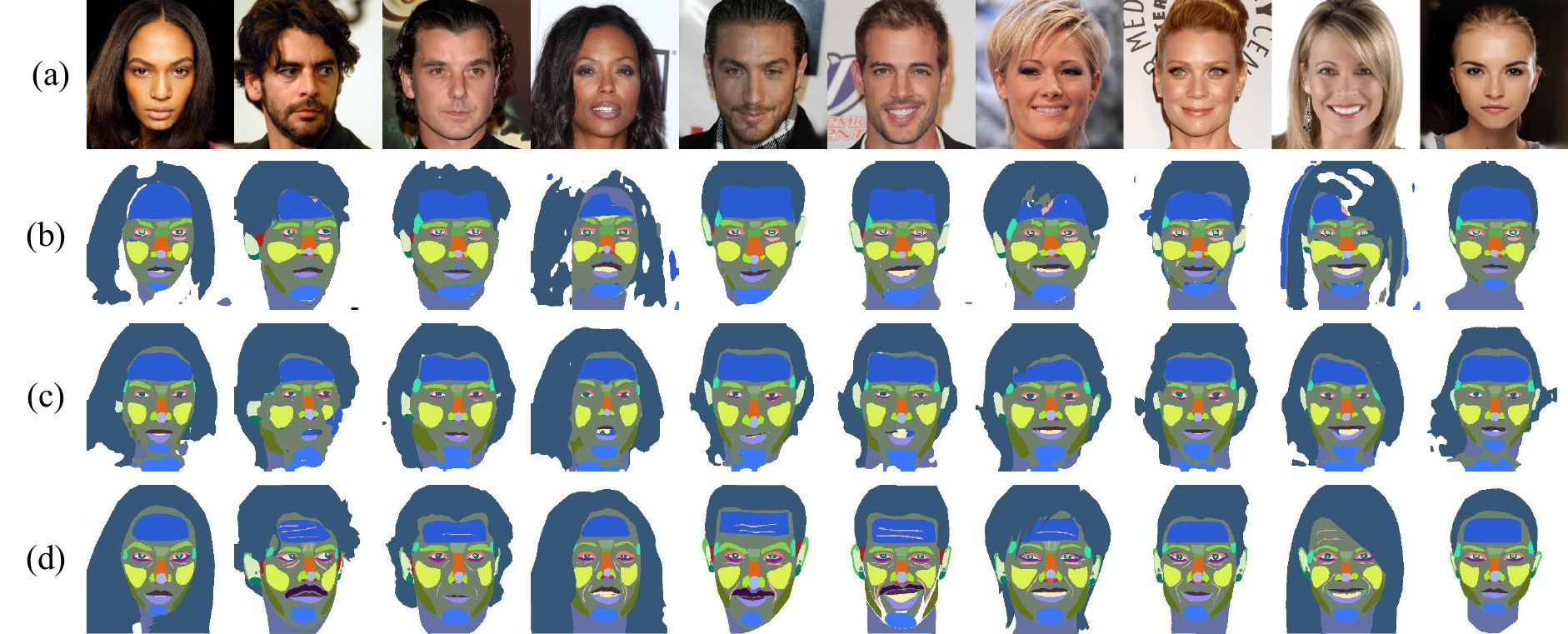}
  \caption{
  \textbf{Visualization of the 1-shot segmenter prediction.}
  Starting from the top row,
  we show the 
  (a) The test set input image,
  (b) segmentation prediction from DatasetGAN 1-shot segmenter,
  (c) segmentation prediction from \ECCVMETHODNAME{}1-shot segmenters
  and (d) the ground truth segmentation annotation.
  } \label{fig:seg:fewshot-comparison}
\end{figure*}

\noindent
\textbf{Evaluation Settings.} 
We compare \ECCVMETHODNAME{} with DatasetGAN~\citep{Zhang2021DatasetGANEL} and CoordGAN~\citep{mu2022coordgan}, 
which are respectively the state-of-the-art 2D GAN-based few-shot segmentation method and the concurrent work on establishing 2D correspondences via 2D GANs.
We also include the 2D representation learning baselines from CoordGAN for reference.

Since \ECCVMETHODNAME{} does not directly accept real images as input,
following~\citet{Zhang2021DatasetGANEL}, 
we sample $10,000$ image-annotation pairs from pretrained GANs as a dataset and train a segmentation model, as shown in Fig.~\ref{fig:seg:3d-consistent-synthetic}.
We evaluate the trained segmentation model on the official DatasetGAN test set to quantify the segmentation accuracy, 
which consists of 16 fine-grained annotated $1024^2$ real-world images. 
For a fair comparison,
we also train an 1-shot DeepLabV3 model
as the baseline,
which uses one annotated pair as the sampling source (Fig.~\ref{fig:seg-1shot-gt}(a)) and follows the data generation pipeline of~\citet{Zhang2021DatasetGANEL}.
Since ~\citet{Zhang2021DatasetGANEL} adopts pretrained StyleGAN under $512^2$ resolution on the FFHQ dataset
while our pretrained GANs are trained over $128^2$ resolution CelebA dataset,
all the test images are bilinear interpolated with a resolution of $256^2$ for evaluation.
We adopt the official implementation of DatasetGAN for data generation and use the default settings for all the segmentation models training.
The standard mIOU is adopted as the segmentation evaluation metric. 

\noindent
\textbf{Results.}
We show the quantitative results the test set of~\citet{Zhang2021DatasetGANEL} in Tab.~\ref{tab:seg:one-shot-segmenter-quant}.
As can be seen, our method achieves comparable performance with the baseline,
and even performs better over some classes like hair and nose,
indicating that our learned correspondences are accurate and smooth.
%
We show the qualitative results in Fig.~\ref{fig:seg:fewshot-comparison}. 
As can be observed, without relying on explicit segmentation supervision, our method can perform 3D consistent segmentation transfer, which is not possible with existing 2D correspondence baselines such as DatasetGAN 1-shot segmenter. This is made possible by establishing plausible correspondence between different semantic regions across NeRFs, despite their structure variations in 3D space.

\begin{table*}[h!]
\centering
\caption{\textbf{mIOU scores of two 1-shot segmenters on DatasetGAN~\citep{Zhang2021DatasetGANEL} test set.} 
The corresponding segmenters are trained over the synthetic dataset generated by two methods.
We show the performance of two versions of \ECCVMETHODNAME{} based on two generators pretrained on different datasets.
The 1-shot segmenter trained on our dataset
is competitive against the counterpart which is trained in high-resolution images, 
demonstrating the merit of the learned correspondence.
}
\label{tab:seg:one-shot-segmenter-quant}
\resizebox{\columnwidth*1}{!}{%
\begin{tabular}{cccccccccccc}
\bottomrule
Method & Mean IOU & Eyes & Mouth & Nose & Cheek & Chin & Hair & Eyebrows & Ears & Jaw & BG   \\ \hline
DatasetGAN 1-shot segmenter & \textbf{56.9} & 40.5 & \textbf{62.6} & 52.5 & 61.6 & 65.5 & \textbf{72.4} & \textbf{59.6} & \textbf{49.0} & 16.4 & \textbf{81.4} \\
\ECCVMETHODNAME{}~1-shot segmenter & 54.6 & \textbf{51.0} & 53.2 & \textbf{55.4} & \textbf{69.2} & \textbf{82.6} & 67.4  & {54.2}    & 40.9 & \textbf{66.9} & 75.06 \\
 \bottomrule
\end{tabular}}
\end{table*}

For further comparison with concurrent work that distills correspondences from 2D GANs,
Tab.~\ref{table:coordgan-seg-propagation} presents the mIOU scores over two real-world datasets.
Following~\cite{mu2022coordgan}, we train an encoder that predicts the source NeRF code $\code_s$ using the techniques described in Sec.~\ref{sec:method:encoder_inversion} and conducts feed-forward inference 
over the input images. 
Our method outperforms 2D learning-based models on this task
and achieves competitive performance compared with CoordGAN,
with the merit of establishing dense correspondences in 3D space.
Compared with building correspondences over 2D pixels,
establishing correspondences in implicit 3D space is exponentially harder, as explained in Sec.~\ref{sec:intro}.
Moreover,
compared with mature 2D GAN families and toolboxes,
the development of 3D GANs is still in its early stage.
Equipping \ECCVMETHODNAME{}~with more developed 3D GANs, \ie~\citet{Chan2022}, can potentially close the gap.

\begin{table*}[t]
  \centering
    \caption{\textbf{IOU comparison for segmentation label propagation}. 
      Our method achieves comparative performance with the 2D representation learning method,
      and is the only method that supports 3D dense correspondence searching over implicit functions.
      $\ast$ means 3D aware.}
    \label{table:coordgan-seg-propagation}
    \begin{tabular}{cccc} 
    \toprule
    Method & CelebA-HQ & DGAN-face  \\
    \midrule
    Swap AE~\citep{park2020swapping} & 24.73 & 5.48  \\
    MoCo~\citep{he2019moco}  & 36.19 & 10.00 \\
    VFS~\citep{xu2021rethinking}  & 38.10 & 8.55 \\
    ResNet50~\citep{he2016deep}  & 39.48 & 11.05 \\
    $\ECCVMETHODNAME{}^{\ast}$ & 45.32 & 19.18 \\
    Pix2Style2Pix~\citep{richardson2020encoding} & 48.50 & 20.36  \\
    CoordGAN~\citep{mu2022coordgan} & 52.25 & 23.78 \\
    \bottomrule
    \end{tabular}
\end{table*}

\subsection{Quantitative Results on Keypoints Transfer}
\label{exp:subsec:quant_result_keypoint}
\begin{table*}[tp]
  \centering
  \caption{\textbf{PCK-Transfer on facial landmarks}. 
  Our method achieves better performance compared to both model-based (row 2) and learning-based (rows 3,4,5) methods.}
  \label{tab:quantitative_landmarks}
  \begin{tabular}{lcccc}
  \bottomrule
  Methods   & Correspondence Supervision &  PCK@0.05 $\uparrow$ & PCK@0.01 $\uparrow$ & AEPE $\downarrow$ \\ 
  \hline
  SIFT Flow~\citep{Liu2011SIFTFD} & - & 92.7 & 32.9  & 5.22   \\
  PDC-Net~\citep{Truong2021LearningAD}   & matching image pairs       &       89.1 &       26.8 & 6.28  \\
  GoCor~\citep{Truong2020GOCorBG}     & matching image pairs       &          87.9 &       24.8 & 6.24  \\ 
  GLU-Net~\citep{Truong2020GLUNetGU}   & matching image pairs       &          90.0 &       30.4& 5.78   \\
  Ours      & GAN-supervised             &         \textbf{95.0}&        \textbf{41.6}  &         \textbf{4.47} \\
  \bottomrule
  \vspace{-2em}
  \end{tabular}
  \end{table*}

\begin{figure*}[tp] 
  \centering
  \begin{minipage}{.49\textwidth}
    \centering
  \includegraphics[width=.99\linewidth, ]{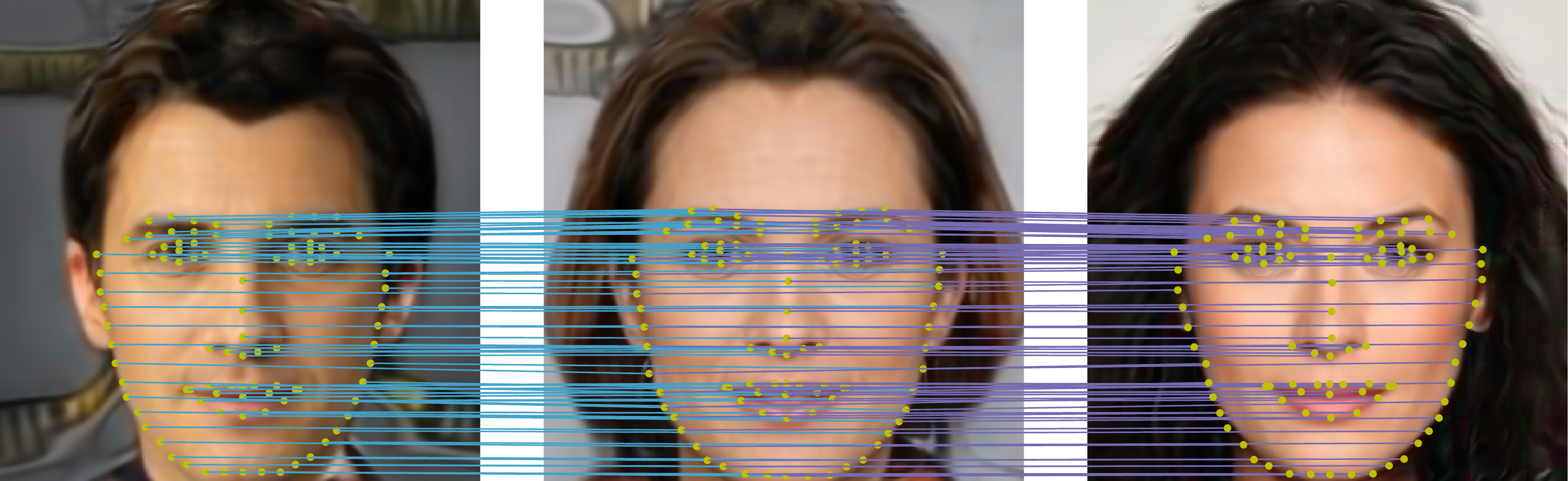}
\end{minipage}%
\begin{minipage}{0.49\textwidth}
    \centering
  \includegraphics[width=0.99\linewidth, ]{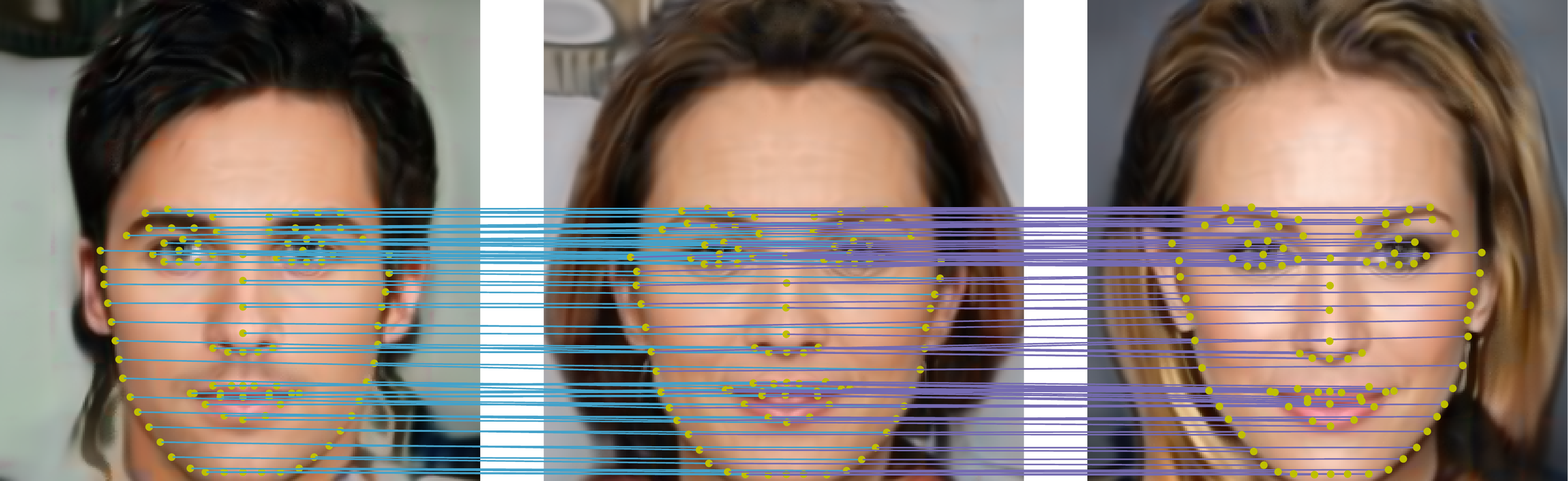}
\end{minipage}
  \caption{
  \textbf{Visualization of learned correspondences via landmark transfer.}
  For each triplet,
  we first predict 98 facial landmarks of the first column acquired through an off-the-shelf model~\citep{Wang_2019_ICCV}.
  We deform the predicted landmarks to the template through network $B$,
  and then further deform the landmarks on the template to another sampled face through network $F$.
  We sample 3D points near the surface of one NeRF and calculate the 
  dense correspondence point on the target NeRF with our deformation network.
  Please zoom in for details.} 
  \label{fig:dense_correspondence_landmarks}
\end{figure*}
Though segmentation label propagation is an intuitive and well-adopted surrogate metric for evaluating learned correspondence, 
we argue that quantitative evaluation using this metric alone is contrived.
Specifically, segmentation label propagation is essentially a pixel-wise classification task, 
which means any errors in dense correspondences within a segment will not be detected.
Moreover, only network $B$ is used in the segmentation label propagation experiment, 
which could not quantitatively evaluate the forward deformation field $F$ in our method.
Therefore, 
we further evaluate our method via keypoints transfer~\citep{zheng2021deep},
which is a regression task with independent ground truth for each transferred landmark.
In our context, this task can be viewed as few-shot 3D facial landmark transfer learning with 1 sample as training data.

For this task, as in the segmentation label propagation pipeline, 
we first use an off-the-shelf facial landmarks prediction model~\citep{Wang_2019_ICCV} to label the template frontal view image with 98 landmarks,
which can be seen in the middle of Fig.~\ref{fig:dense_correspondence_landmarks}.
Since these points are in the image space, 
we first unproject them back to the template NeRF 3D space by appending the corresponding depth values viewing these landmarks positions,
which we denote as $\gP_{0}^{lms}=\{\point_{0}^{lms(k)}\}_{k=1}^{98}$.
After that, we resort to $F$ model and deform these unprojected 3D points to $\nerf_t$ by $F(\point_0^{lms(*)}, \code_t)=\point_t^{lms(*)}$.
The deformed points $\point_t^{lms(*)}$ are projected back to the image space as the transferred 2D facial landmarks of $\nerf_t$.

\noindent
\textbf{Evaluation Settings}
We also compare our method with both 3D model-based method (3DMM~\citep{blanz1999morphable}) as well as current state-of-the-art 2D learning-based matching method~\citep{Truong2020GLUNetGU,Truong2020GOCorBG,Truong2021LearningAD} as our baselines.
We regard the output of a representative landmark detector 
MTCNN~\citep{Zhang2016JointFD} as ground-truth.
For baselines,
we also consider the hand-crated descriptor SIFT Flow~\citep{Liu2011SIFTFD}
as well as several learned descriptors~\citep{Truong2020GOCorBG,Truong2021LearningAD,Truong2020GLUNetGU} 
that attain state-of-the-art performance on commonly used 
dense correspondence benchmarks (\eg, MegaDepth~\citep{Li2018MegaDepthLS}).
For the baselines, 
we use the officially released models to conduct inference in our experiments.
We employ the Percentage of Correctly Keypoints (PCK) and Average End Point Error (AEPE) as the evaluation metrics.

\noindent
\textbf{Results.}
We evaluate the performance over
$5,000$ randomly sampled human faces with different view angles and show the quantitative results in Tab.~\ref{tab:quantitative_landmarks}.
At threshold PCK@0.01, \ECCVMETHODNAME{}~achieves
41.6 against competitive baselines.
This result strongly supports the effectiveness of \ECCVMETHODNAME{}.

\subsection{Ablation Study}
\noindent
\begin{figure*}[h!]
    \centering 
    \includegraphics[width=1\textwidth]{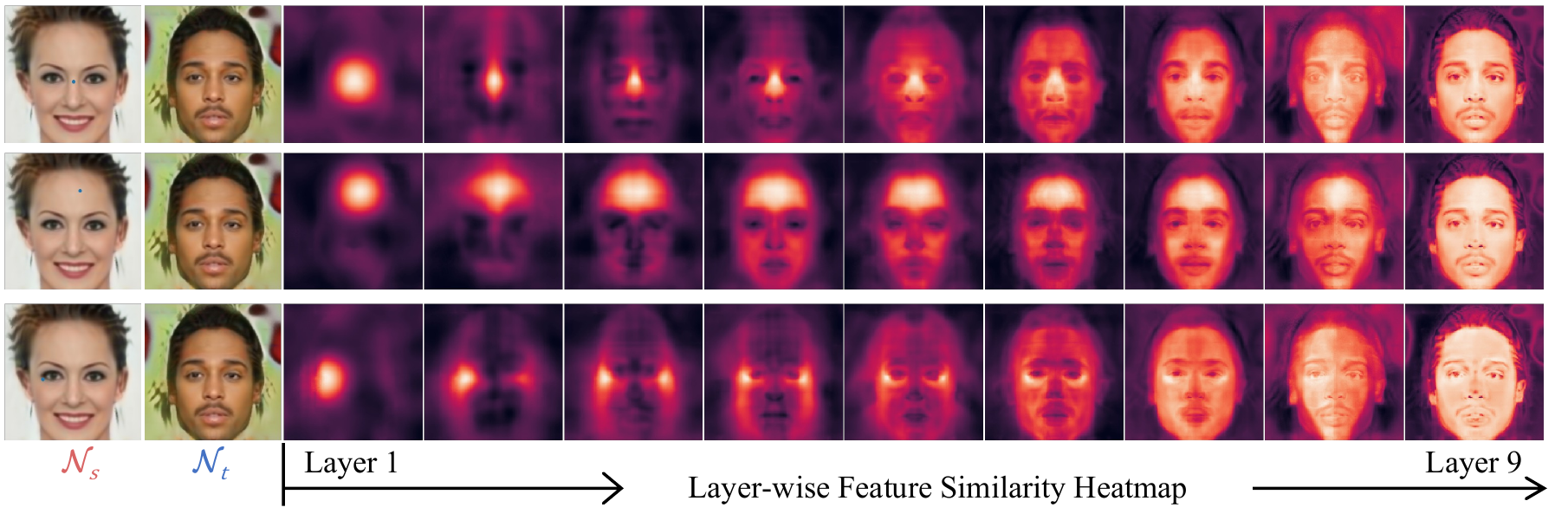}
\vspace{-0.9cm}
\caption{
\textbf{Layer-wise feature correlation between projected features of two NeRF.} 
The NeRFs are sampled from a pretrained $\pi$-GAN generator from \emph{shallow (leftmost)} to \emph{deep (rightmost)}.
Since $\pi-$GAN generator adopts an 8-layer MLP design appended with a view-dependent MLP layer, 
here from left to right, we show the feature similarity heatmap from the 1st layer (3rd column) to the 9th layer (last column) of the pretrained generator.
We project the features of 3D NeRF to 2D using Eq.~\ref{eq:volume_render} for better visualization.
In each row, a random 2D point from the source NeRF is selected to calculate layer-wise feature similarity heatmaps with the projected feature map of the target NeRF. 
}
\label{fig:3dae-abla-featanalysis}
\end{figure*}

\noindent
\textbf{Selection of Generator Feature.}
\label{sec:exp:abla:feature_analysis}
In our work, we select multiple layers from the generator as the training supervisions 
of the feature similiary losses depicted in Eq.~\ref{eq:L_similarity_B} and~\ref{eq:L_similarity_F}.
Here we justify the intuition behind this.
%
%
Different from the feed-forward models~\citep{Simonyan2015VeryDC,he2016deep},
generative models like GANs is trained to decode information from a compact latent code.
Therefore, features from earlier layers should contain more high-level semantics information while later layers contain more instance-specific texture information.
To justify this intuition,
we show the layer-wise feature similarity heatmap between the projected 2D feature maps of $\nerf_s$ and $\nerf_t$
over a pretrained $\pi$-GAN generator
in Fig.~\ref{fig:3dae-abla-featanalysis}.
Specifically, 
given $\nerf_s$ and $\nerf_t$,
we calculate the 2D feature maps by integrating the features of points along each rays using the volume rendering equation depicted in~Eq.~\ref{eq:volume_render} and get two sets of features maps $F_{s} = \gR^{N*H*W*C}$ and $F_{t}=\gR^{N*H*W*C}$, where $N=9$ is the layer number of $\pi$-GAN generators.
Given a 2D coordinate $(u,v)$,
we retrieve its corresponding features $F^{u,v}_{s} \in \gR^{N*C}$ from the source feature maps $F_{s}$ and calculate the cosine similarity with the target feature maps $F_{t}$ within each layer.

As can be seen, the generator features from different layers encode semantics from different levels, where the semantics compactness linearly decreases as the network goes deeper.
Surprisingly, the early generator features are even robust under the symmetric semantics such as the right and left corners of the eyes (3rd row).
This is an indispensable property in establishing dense correspondences where a 3D point from the source left eye should not establish correspondence to points in the right eye region of the target.
Thus, we choose the normalized features from the first $5$ layers as the supervision signals of the \ECCVMETHODNAME{}, which encode unambiguous correspondence information.
This property has also been validated in 2D generative models~\citep{peebles2022gansupervised,yang2022Unsupervised,StyleGAN3D},
where different layers of pretrained StyleGAN encode different types of information.

\begin{figure*}[t]
        \centering
        \includegraphics[width=\textwidth]{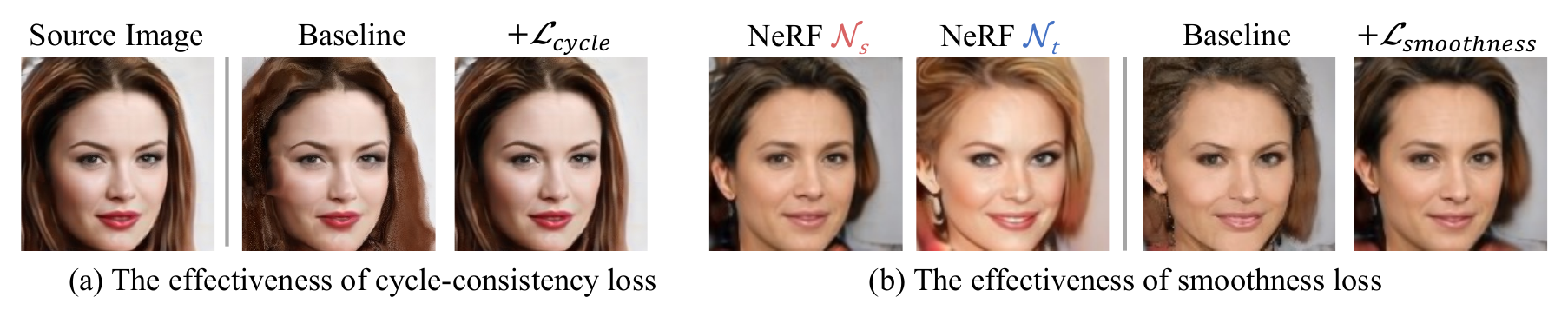}
        \caption{\textbf{(a) Rendering from the self-reconstructed point through cycle deformation.}
        From the left is the input image, reconstructed with and without cycle-consistency loss. 
        The deformation model trained with cycle-consistency loss can perfectly reconstruct itself, while the one without cycle-consistency loss leads to distortions. 
        \textbf{(b) Output from deformation network trained without and with deformation smoothness loss.} 
        We demonstrate the effectiveness of $L_smooth$ via texture transfer from $A$ to $B$.
        The lack of smoothness regularization leads to distorted visual results. Better zoom in for a better experience.}
        \label{fig:abla_loss}
\end{figure*}

\begin{figure*}[h!] 
  \centering
  \includegraphics[width=1\linewidth, ]{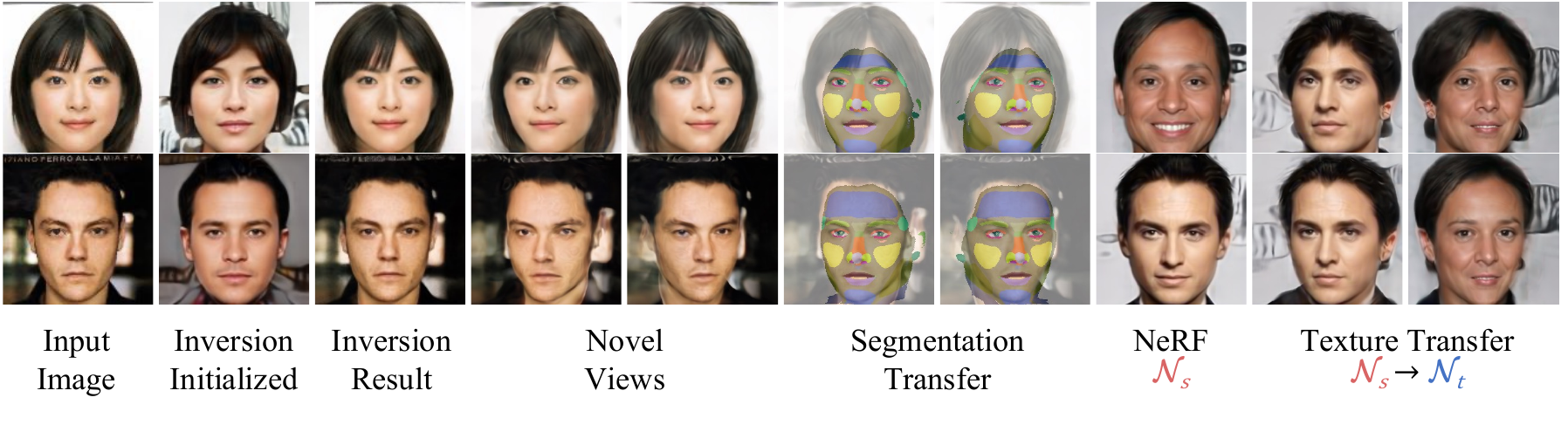}
  \vspace{-0.5cm}
  \caption{
  \textbf{Extending \ECCVMETHODNAME{} to real images.}
  To apply \ECCVMETHODNAME{} to real-world image,
  we first inverse the real-world images (1st column) into the latent space of the 3D GAN (2nd and 3rd columns).
  Beyond novel view synthesis (4th and 5th columns),
  \ECCVMETHODNAME{}~also supports 3D consistent segmentation transfer (6th and 7th columns).
  Given the reference NeRFs (8th column),
  our method could edit the texture of given identities without changing the overall shape.}
  \label{fig:application:real-case-transfer}
\end{figure*}
\noindent
\textbf{Deformation Regularization Terms.}
We validate the efficacy of our regularization terms in terms of qualitative results,
including the cycle consistency term and deformation smoothness term 
%
%
To construct a baseline for evaluation,
we remove the correspondence deformation smoothness loss term
and only apply supervision from the feature similarity loss on network training.
To evaluate the effect of cycle consistency regularization,
we train a baseline without cycle consistency loss term and visualize the self-reconstruction as well as texture transfer results using the trained dual deformation field.
As shown in Fig. \ref{fig:abla_loss}(a),
cycle-consistency term encourages the consistency property, 
that after the forward and backward deformation a point from the source shape will map to itself.
Meanwhile,
the model trained with cycle consistency term learns less noisy deformation, 
which is essential when conducting downstream tasks such as texture transfer.
As shown in Fig. \ref{fig:abla_loss}(b), 
without the deformation smoothness regularization,
the network tends to learn noisy deformation which leads to distortions in the final rendering.
For the second order cycle consistency regularization,
we find it has similar effect with $\gL_{cycle}$ in qualitative performance.
Moreover,
we set $\lambda_{cycle}^{2nd}=0$ and conduct the segmentation transfer evaluation as in Tab.~\ref{tab:seg:one-shot-segmenter-quant} and observe a mIOU degrade from $56.9$ to $55.3$,
which validates $\gL_{cycle}^{2nd}$ could improve the deformation field performance.

\noindent
\textbf{Curriculum Training.}
To show that our proposed curriculum training strategy could help regularize the training and facilitate convergence,
we report the values of feature similarity loss over the evaluation set,
with different curriculum steps adopted during training. 
Using $16/128/1024/4096$ steps the 
loss are respectively $0.455/0.410/0.310/0.287$,
which demonstrates the effectiveness of our method.
\subsection{Extending \ECCVMETHODNAME{} to Real Images}

\begin{figure*}[h!] 
  \centering
  \includegraphics[width=1\linewidth, ]{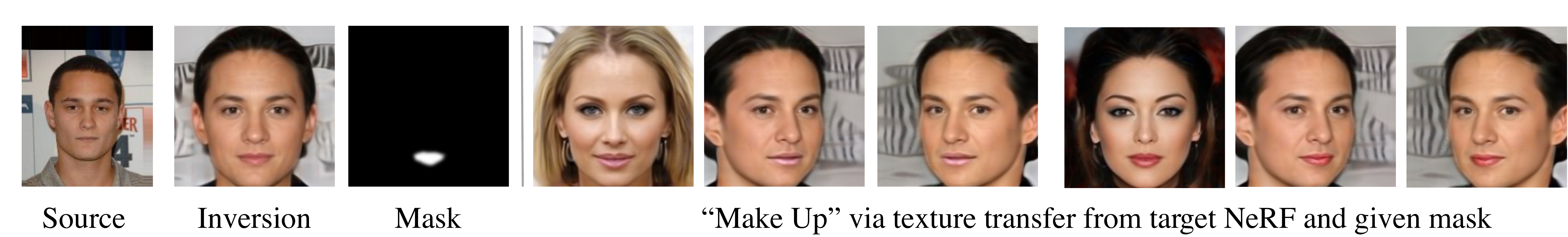}
  \vspace{-0.5cm}
  \caption{
  \textbf{Mask-guided texture transfer over real cases. }
  In the second column, we show the projected image of the GAN inversion of the source image. 
  For mask-guided texture manipulation, we sample two synthetic NeRF from the pretrained GAN (4th and 7th columns) and conduct texture transfer guided by the foreground mask. 
}
\label{fig:application:real-case-lipstick}
\end{figure*}

\noindent\textbf{Training An Encoder for Inversion.}\label{sec:method:encoder_inversion}
To apply \ECCVMETHODNAME{}~real-world images,
we use two encoders, namely an encoder denoted by $E(\cdot, \theta_{G})$ to invert the input image to the latent space of the NeRF-based GAN and another encoder represented by $E(\cdot, \theta_{\ECCVMETHODNAME{}})$ to invert the input image to the deformation conditions.
%
Specifically,
rather than directly output the low-dimensional deformation code $\code$, 
here we follow the observations of~\citet{tov2021designing} which project the $\mathcal{Z}$-space code $\code$ to the $\mathcal{W}{+}$ space for better performance.
Since the NeRF-based GAN (\ie, $\pi$-GAN) already follows this design,
here we further augment each of the deformation fields with a mapping function $\gM_\ECCVMETHODNAME{}$~\citep{chan2020pi,sitzmann2020siren}.
During inversion, the corresponding encoders directly output the $\gW+$ space modulations, \ie,
$E(I, \theta_{G})=\bm\beta_{G}^{I}, \bm\gamma_{G}^{I}$ and $E(I, \theta_{\ECCVMETHODNAME{}})=\bm\beta_{\ECCVMETHODNAME{}}^{I}, \bm\gamma_{\ECCVMETHODNAME{}}^{I}$. 

The encoders are trained in two stages.
In the first stage,
we train the encoder $E(\cdot, \theta_{G})$ where the output latent codes $\bm\beta_{G}^{I}, \bm\gamma_{G}^{I}$ are fed into the NeRF-based GAN to render a replicate of the images $\hat{I} = G(\bm\beta_{G}^{I}, \bm\gamma_{G}^{I}, \bm\xi_{I})$,
where $\bm\xi_{I}$ is the estimated camera pose of the input image using an off-the-shelf pose estimator.
After $E(\cdot, \theta_{G})$ converges,
in the second stage we train the encoder $E(\cdot, \theta_{D})$ to output the corresponding deformations conditions $\bm\beta_{B}^{I}, \bm\gamma_{B}^{I}$ and $\bm\beta_{F}^{I}, \bm\gamma_{F}^{I}$.
Given the inverted latent code $\bm\beta_{G}^{I}, \bm\gamma_{G}^{I}$ of the input image,
we conduct self texture transfer described in Sec.~\ref{exp:subsec:texture_transfer} to replicate the input image.
%
To stabilize training,
we also include synthetic samples from the pre-trained NeRF GAN as training data.
Given a latent code $\code{\sim}p_{\code}$,   
we get the paired modulations $\bm\beta_{\code}, \bm\gamma_{\code}$ and synthesized image $I_{\code}=G(\code, \bm\xi)$ under a random camera pose $\bm\xi{\sim}p_{\bm\xi}$ as training samples.
Apart from image reconstruction loss, the predicted modulations from $E(\cdot, \theta_{D})$ are encouraged to mimic the synthetic ground truth.
We find the synthetic latent code regularization could stabilize the deformation encoder training.
Following ~\citet{tov2021designing}, 
the encoders predict the offsets of the mean modulations of the corresponding mapping function
$\bm\beta_{0}^{\ECCVMETHODNAME{}}, \bm\gamma_{0}^{\ECCVMETHODNAME{}}$
for better initialization.
The overall training objectives are:
1) \emph{Image Reconstruction Loss}:
We utilize the pixel-wise $\mathcal{L}_{\text{2}}$ as well as the LPIPS loss  $\mathcal{L}_{\text{LPIPS}}$~\citep{zhang2018perceptual} as the image reconstruction supervisions:
\begin{equation}\label{method:encoder:img_rec_reg}
\begin{split}
    \mathcal{L}_{image} &= \mathcal{L}_{\text{2}}(I, G(E(I, \theta_{G}),\bm\xi_{I})) \\
    &+ \mathcal{L}_{\text{LPIPS}}(I, G(E(I, \theta_{G}),\bm\xi_{I})),
\end{split}
\end{equation}
2) \emph{Latent Codes Regularization}:
We regularize the encoded latent codes to match the pseudo ground truth latent codes distribution:
\begin{equation}\label{method:encoder:latent_reg}
    \mathcal{L}_{latent} = \mathcal{L}_{2}((\bm\beta_{\vz}, \bm\gamma_{\vz}), E(I_\vz, \theta_{D})).
\end{equation}


\noindent
\textbf{Results.}
Here we show the texture transfer results over real images in Fig.~\ref{fig:application:real-case-transfer}. 
As can be seen,
our hybrid inversion method could faithfully reconstruct the given real images without affecting the view synthesis ability of NeRF.
Furthermore, with \ECCVMETHODNAME{},
accurate 3D-consistent segmentation transfer and faithful texture transfer become possible,
which is beyond the reach of existing 2D methods.
We further show a mask-guided texture transfer applied over real cases in Fig.~\ref{fig:application:real-case-lipstick},
which shows the potential of our method over real-world applications beyond basic texture transfer and segmentation labeling.


\section{Conclusion}
\label{sec:conclusion}
%
In this work we propose to leverage a pre-trained NeRF-based GAN, $\pi$-GAN in our case,
to build dense correspondence between NeRF representations of different objects 
within the same category.
The key insight is that 
the pre-trained GAN possesses three important properties that can help alleviate the challenges of this task,
namely 1) instance-specific latent codes that holistically capture the global structure of different NeRFs,
2) geometric-aware generator features that reflect local geometric details of different NeRFs,
and 3) the manifold of NeRFs that serves as a source of infinite NeRF samples.
Based on the three properties,
we respectively propose a generalizable model, 
referred to as Dual Deformation Field,
a learning objective based on generator features that approximate geometric distances in feature space,
and finally an effective curriculum training strategy that feeds samples with growing complexity.
To the best of our knowledge,
this is the first method that tries to establish dense correspondence across NeRF representations.
Our experiments demonstrate that dense correspondences between NeRFs learned from our framework are accurate, 
smooth, and robust,
making them applicable in various downstream applications.
\newline
\newline
\noindent
\textbf{Data Availability.}
The datasets that support the ﬁndings of this study are all publicly available for the research purpose.
    
    \newpage
    


    %
    %

    \bibliographystyle{spbasic}      
    \bibliography{bibs/naming/short.bib,bibs/bibdesk.bib}

\begin{thebibliography}{84}
\providecommand{\natexlab}[1]{#1}
\providecommand{\url}[1]{{#1}}
\providecommand{\urlprefix}{URL }
\expandafter\ifx\csname urlstyle\endcsname\relax
  \providecommand{\doi}[1]{DOI~\discretionary{}{}{}#1}\else
  \providecommand{\doi}{DOI~\discretionary{}{}{}\begingroup
  \urlstyle{rm}\Url}\fi
\providecommand{\eprint}[2][]{\url{#2}}

\bibitem[{Bau et~al.(2020)Bau, Zhu, Strobelt, Lapedriza, Zhou, and
  Torralba}]{bau2020units}
Bau D, Zhu JY, Strobelt H, Lapedriza A, Zhou B, Torralba A (2020) Understanding
  the role of individual units in a deep neural network. PNAS
  \doi{10.1073/pnas.1907375117},
  \urlprefix\url{https://www.pnas.org/content/early/2020/08/31/1907375117}

\bibitem[{Blanz and Vetter(1999)}]{blanz1999morphable}
Blanz V, Vetter T (1999) A morphable model for the synthesis of 3d faces. In:
  SIGGRAPH

\bibitem[{Brock et~al.(2019)Brock, Donahue, and Simonyan}]{Brock2019LargeSG}
Brock A, Donahue J, Simonyan K (2019) Large scale {GAN} training for high
  fidelity natural image synthesis. In: ICLR, OpenReview.net,
  \urlprefix\url{https://openreview.net/forum?id=B1xsqj09Fm}

\bibitem[{Chan et~al.(2021{\natexlab{a}})Chan, Monteiro, Kellnhofer, Wu, and
  Wetzstein}]{Chan2021piGANPI}
Chan E, Monteiro M, Kellnhofer P, Wu J, Wetzstein G (2021{\natexlab{a}})
  pi-gan: Periodic implicit generative adversarial networks for 3d-aware image
  synthesis. In: CVPR

\bibitem[{Chan et~al.(2021{\natexlab{b}})Chan, Lin, Chan, Nagano, Pan, Mello,
  Gallo, Guibas, Tremblay, Khamis, Karras, and Wetzstein}]{Chan2021}
Chan ER, Lin CZ, Chan MA, Nagano K, Pan B, Mello SD, Gallo O, Guibas L,
  Tremblay J, Khamis S, Karras T, Wetzstein G (2021{\natexlab{b}}) Efficient
  geometry-aware {3D} generative adversarial networks. In: arXiv

\bibitem[{Chan et~al.(2021{\natexlab{c}})Chan, Monteiro, Kellnhofer, Wu, and
  Wetzstein}]{chan2020pi}
Chan ER, Monteiro M, Kellnhofer P, Wu J, Wetzstein G (2021{\natexlab{c}})
  {pi-GAN}: {P}eriodic implicit generative adversarial networks for {3D}-aware
  image synthesis. In: CVPR

\bibitem[{Chan et~al.(2022)Chan, Lin, Chan, Nagano, Pan, Mello, Gallo, Guibas,
  Tremblay, Khamis, Karras, and Wetzstein}]{Chan2022}
Chan ER, Lin CZ, Chan MA, Nagano K, Pan B, Mello SD, Gallo O, Guibas L,
  Tremblay J, Khamis S, Karras T, Wetzstein G (2022) Efficient geometry-aware
  {3D} generative adversarial networks. In: CVPR

\bibitem[{Chang et~al.(2015)Chang, Funkhouser, Guibas, Hanrahan, Huang, Li,
  Savarese, Savva, Song, Su, Xiao, Yi, and Yu}]{shapenet2015}
Chang AX, Funkhouser T, Guibas L, Hanrahan P, Huang Q, Li Z, Savarese S, Savva
  M, Song S, Su H, Xiao J, Yi L, Yu F (2015) {ShapeNet: An Information-Rich 3D
  Model Repository}. Tech. Rep. arXiv:1512.03012 [cs.GR], Stanford University
  --- Princeton University --- Toyota Technological Institute at Chicago

\bibitem[{Chen et~al.(2017)Chen, Papandreou, Schroff, and
  Adam}]{Chen2017RethinkingAC}
Chen LC, Papandreou G, Schroff F, Adam H (2017) Rethinking atrous convolution
  for semantic image segmentation. arXiv abs/1706.05587

\bibitem[{Chen and Zhang(2019)}]{Chen2019LearningIF}
Chen Z, Zhang H (2019) Learning implicit fields for generative shape modeling.
  In: CVPR, pp 5932--5941

\bibitem[{Deng et~al.(2021)Deng, Yang, and Tong}]{deng2021deformed}
Deng Y, Yang J, Tong X (2021) Deformed implicit field: Modeling 3d shapes with
  learned dense correspondence. In: CVPR, pp 10286--10296

\bibitem[{Dosovitskiy et~al.(2017)Dosovitskiy, Ros, Codevilla, Lopez, and
  Koltun}]{DBLP:journals/corr/abs-1711-03938}
Dosovitskiy A, Ros G, Codevilla F, Lopez A, Koltun V (2017) Carla: An open
  urban driving simulator. In: Proc. CoRL

\bibitem[{Dumoulin et~al.(2018)Dumoulin, Perez, Schucher, Strub, Vries,
  Courville, and Bengio}]{dumoulin2018feature-wise}
Dumoulin V, Perez E, Schucher N, Strub F, Vries Hd, Courville A, Bengio Y
  (2018) Feature-wise transformations. Distill \doi{10.23915/distill.00011},
  https://distill.pub/2018/feature-wise-transformations

\bibitem[{Egger et~al.(2020)Egger, Smith, Tewari, Wuhrer, Zollh{\"o}fer,
  Beeler, Bernard, Bolkart, Kortylewski, Romdhani, Theobalt, Blanz, and
  Vetter}]{Egger20203DMF}
Egger B, Smith W, Tewari A, Wuhrer S, Zollh{\"o}fer M, Beeler T, Bernard F,
  Bolkart T, Kortylewski A, Romdhani S, Theobalt C, Blanz V, Vetter T (2020) 3d
  morphable face models---past, present, and future. TOG 39:1 -- 38

\bibitem[{Eisenberger et~al.(2021)Eisenberger, Novotn{\'y}, Kerchenbaum,
  Labatut, Neverova, Cremers, and Vedaldi}]{Eisenberger2021NeuroMorphUS}
Eisenberger M, Novotn{\'y} D, Kerchenbaum G, Labatut P, Neverova N, Cremers D,
  Vedaldi A (2021) Neuromorph: Unsupervised shape interpolation and
  correspondence in one go. In: CVPR

\bibitem[{Eslami et~al.(2018)Eslami, Rezende, Besse, Viola, Morcos, Garnelo,
  Ruderman, Rusu, Danihelka, Gregor, Reichert, Buesing, Weber, Vinyals,
  Rosenbaum, Rabinowitz, King, Hillier, Botvinick, Wierstra, Kavukcuoglu, and
  Hassabis}]{Eslami2018NeuralSR}
Eslami SMA, Rezende DJ, Besse F, Viola F, Morcos AS, Garnelo M, Ruderman A,
  Rusu AA, Danihelka I, Gregor K, Reichert DP, Buesing L, Weber T, Vinyals O,
  Rosenbaum D, Rabinowitz NC, King H, Hillier C, Botvinick MM, Wierstra D,
  Kavukcuoglu K, Hassabis D (2018) Neural scene representation and rendering.
  Science 360:1204 -- 1210

\bibitem[{Fan et~al.(2019)Fan, Hu, Chen, and Peng}]{Fan2019BoostingLS}
Fan Z, Hu X, Chen C, Peng S (2019) Boosting local shape matching for dense 3d
  face correspondence. In: CVPR, pp 10936--10946

\bibitem[{Gafni et~al.(2021)Gafni, Thies, Zollh{\"o}fer, and
  Nie{\ss}ner}]{Gafni_2021_CVPR}
Gafni G, Thies J, Zollh{\"o}fer M, Nie{\ss}ner M (2021) Dynamic neural radiance
  fields for monocular 4d facial avatar reconstruction. In: CVPR, pp 8649--8658

\bibitem[{Goodfellow et~al.(2014)Goodfellow, Pouget-Abadie, Mirza, Xu,
  Warde-Farley, Ozair, Courville, and Bengio}]{Goodfellow2014GenerativeAN}
Goodfellow IJ, Pouget-Abadie J, Mirza M, Xu B, Warde-Farley D, Ozair S,
  Courville AC, Bengio Y (2014) Generative adversarial nets. In: NIPS

\bibitem[{Gu et~al.(2022)Gu, Liu, Wang, and Theobalt}]{gu2022stylenerf}
Gu J, Liu L, Wang P, Theobalt C (2022) Stylenerf: A style-based 3d aware
  generator for high-resolution image synthesis. In: ICLR,
  \urlprefix\url{https://openreview.net/forum?id=iUuzzTMUw9K}

\bibitem[{Guo et~al.(2021)Guo, Chen, Liang, Liu, Bao, and
  Zhang}]{guo2021adnerf}
Guo Y, Chen K, Liang S, Liu Y, Bao H, Zhang J (2021) Ad-nerf: Audio driven
  neural radiance fields for talking head synthesis. In: ICCV

\bibitem[{Halimi et~al.(2019)Halimi, Litany, Rodola, Bronstein, and
  Kimmel}]{halimi2019unsupervised}
Halimi O, Litany O, Rodola E, Bronstein AM, Kimmel R (2019) Unsupervised
  learning of dense shape correspondence. In: CVPR, pp 4370--4379

\bibitem[{He et~al.(2016)He, Zhang, Ren, and Sun}]{he2016deep}
He K, Zhang X, Ren S, Sun J (2016) Deep residual learning for image
  recognition. In: CVPR

\bibitem[{He et~al.(2019)He, Fan, Wu, Xie, and Girshick}]{he2019moco}
He K, Fan H, Wu Y, Xie S, Girshick R (2019) Momentum contrast for unsupervised
  visual representation learning. arXiv

\bibitem[{Hong et~al.(2022)Hong, Peng, Xiao, Liu, and Zhang}]{hong2021headnerf}
Hong Y, Peng B, Xiao H, Liu L, Zhang J (2022) Headnerf: A real-time nerf-based
  parametric head model. In: CVPR

\bibitem[{Jahanian et~al.(2020)Jahanian, Chai, and Isola}]{gansteerability}
Jahanian A, Chai L, Isola P (2020) On the "steerability" of generative
  adversarial networks. In: ICLR

\bibitem[{Kaick et~al.(2011)Kaick, Zhang, Hamarneh, and
  Cohen-Or}]{Kaick2011ASO}
Kaick OV, Zhang H, Hamarneh G, Cohen-Or D (2011) A survey on shape
  correspondence. Computer Graphics Forum 30

\bibitem[{Karras et~al.(2019{\natexlab{a}})Karras, Laine, and
  Aila}]{Karras2019ASG}
Karras T, Laine S, Aila T (2019{\natexlab{a}}) A style-based generator
  architecture for generative adversarial networks. In: CVPR, pp 4396--4405

\bibitem[{Karras et~al.(2019{\natexlab{b}})Karras, Laine, and
  Aila}]{karras2019style}
Karras T, Laine S, Aila T (2019{\natexlab{b}}) A style-based generator
  architecture for generative adversarial networks. In: CVPR

\bibitem[{Kingma and Ba(2015)}]{Kingma2015AdamAM}
Kingma DP, Ba J (2015) Adam: A method for stochastic optimization. In: ICLR,
  vol abs/1412.6980

\bibitem[{Li and Snavely(2018)}]{Li2018MegaDepthLS}
Li Z, Snavely N (2018) Megadepth: Learning single-view depth prediction from
  internet photos. In: ICCV, pp 2041--2050

\bibitem[{Li et~al.(2021)Li, Niklaus, Snavely, and Wang}]{li2020neural}
Li Z, Niklaus S, Snavely N, Wang O (2021) Neural scene flow fields for
  space-time view synthesis of dynamic scenes. In: CVPR

\bibitem[{Litany et~al.(2017)Litany, Remez, Rodol{\`a}, Bronstein, and
  Bronstein}]{Litany2017DeepFM}
Litany O, Remez T, Rodol{\`a} E, Bronstein A, Bronstein M (2017) Deep
  functional maps: Structured prediction for dense shape correspondence. In:
  ICCV, pp 5660--5668

\bibitem[{Liu et~al.(2011)Liu, Yuen, and Torralba}]{Liu2011SIFTFD}
Liu C, Yuen J, Torralba A (2011) Sift flow: Dense correspondence across scenes
  and its applications. PAMI 33:978--994

\bibitem[{Liu and
  Liu(2020)}]{learning-implicit-functions-for-topology-varying-dense-3d-shape-correspondence}
Liu F, Liu X (2020) Learning implicit functions for topology-varying dense 3d
  shape correspondence. In: NIPS, Virtual

\bibitem[{Liu et~al.(2015)Liu, Luo, Wang, and Tang}]{liu2015faceattributes}
Liu Z, Luo P, Wang X, Tang X (2015) Deep learning face attributes in the wild.
  In: ICCV

\bibitem[{Loper et~al.(2015)Loper, Mahmood, Romero, Pons-Moll, and
  Black}]{loper2015smpl}
Loper M, Mahmood N, Romero J, Pons-Moll G, Black MJ (2015) Smpl: A skinned
  multi-person linear model. TOG 34(6):1--16

\bibitem[{Mescheder et~al.(2019)Mescheder, Oechsle, Niemeyer, Nowozin, and
  Geiger}]{mescheder2019occupancy}
Mescheder L, Oechsle M, Niemeyer M, Nowozin S, Geiger A (2019) Occupancy
  networks: Learning 3d reconstruction in function space. In: CVPR, pp
  4460--4470

\bibitem[{Mescheder et~al.(2018)Mescheder, Geiger, and
  Nowozin}]{Mescheder2018WhichTM}
Mescheder LM, Geiger A, Nowozin S (2018) Which training methods for gans do
  actually converge? In: ICML

\bibitem[{Mildenhall et~al.(2020)Mildenhall, Srinivasan, Tancik, Barron,
  Ramamoorthi, and Ng}]{mildenhall2020nerf}
Mildenhall B, Srinivasan PP, Tancik M, Barron JT, Ramamoorthi R, Ng R (2020)
  Nerf: Representing scenes as neural radiance fields for view synthesis. In:
  ECCV, Springer, pp 405--421

\bibitem[{Mu et~al.(2022)Mu, De~Mello, Yu, Vasconcelos, Wang, Kautz, and
  Liu}]{mu2022coordgan}
Mu J, De~Mello S, Yu Z, Vasconcelos N, Wang X, Kautz J, Liu S (2022) Coordgan:
  Self-supervised dense correspondences emerge from gans. In: CVPR

\bibitem[{Niemeyer and Geiger(2021)}]{GIRAFFE}
Niemeyer M, Geiger A (2021) Giraffe: Representing scenes as compositional
  generative neural feature fields. In: CVPR

\bibitem[{Noguchi et~al.(2021)Noguchi, Sun, Lin, and Harada}]{2021narf}
Noguchi A, Sun X, Lin S, Harada T (2021) Neural articulated radiance field. In:
  ICCV

\bibitem[{Or-El et~al.(2021)Or-El, Luo, Shan, Shechtman, Park, and
  Kemelmacher-Shlizerman}]{orel2021stylesdf}
Or-El R, Luo X, Shan M, Shechtman E, Park JJ, Kemelmacher-Shlizerman I (2021)
  Style{SDF}: {H}igh-{R}esolution {3D}-{C}onsistent {I}mage and {G}eometry
  {G}eneration. In: CVPR

\bibitem[{Pan et~al.(2021)Pan, Dai, Liu, Loy, and Luo}]{pan_2d_2020}
Pan X, Dai B, Liu Z, Loy CC, Luo P (2021) Do 2d gans know 3d shape?
  unsupervised 3d shape reconstruction from 2d image gans. In: ICLR

\bibitem[{Park et~al.(2019)Park, Florence, Straub, Newcombe, and
  Lovegrove}]{park_deepsdf_2019}
Park JJ, Florence P, Straub J, Newcombe R, Lovegrove S (2019) {DeepSDF}:
  Learning continuous signed distance functions for shape representation. In:
  CVPR, {IEEE}, pp 165--174, \doi{10.1109/CVPR.2019.00025},
  \urlprefix\url{https://ieeexplore.ieee.org/document/8954065/}

\bibitem[{Park et~al.(2021{\natexlab{a}})Park, Sinha, Barron, Bouaziz, Goldman,
  Seitz, and Martin-Brualla}]{park_deformable_2020}
Park K, Sinha U, Barron JT, Bouaziz S, Goldman DB, Seitz SM, Martin-Brualla R
  (2021{\natexlab{a}}) Nerfies: Deformable neural radiance fields. In: ICCV

\bibitem[{Park et~al.(2021{\natexlab{b}})Park, Sinha, Barron, Bouaziz, Goldman,
  Seitz, and Martin-Brualla}]{park2021nerfies}
Park K, Sinha U, Barron JT, Bouaziz S, Goldman DB, Seitz SM, Martin-Brualla R
  (2021{\natexlab{b}}) Nerfies: Deformable neural radiance fields. In: ICCV

\bibitem[{Park et~al.(2021{\natexlab{c}})Park, Sinha, Hedman, Barron, Bouaziz,
  Goldman, Martin-Brualla, and Seitz}]{park2021hypernerf}
Park K, Sinha U, Hedman P, Barron JT, Bouaziz S, Goldman DB, Martin-Brualla R,
  Seitz SM (2021{\natexlab{c}}) Hypernerf: A higher-dimensional representation
  for topologically varying neural radiance fields. TOG 40(6)

\bibitem[{Park et~al.(2020)Park, Zhu, Wang, Lu, Shechtman, Efros, and
  Zhang}]{park2020swapping}
Park T, Zhu JY, Wang O, Lu J, Shechtman E, Efros AA, Zhang R (2020) Swapping
  autoencoder for deep image manipulation. In: NIPS

\bibitem[{Peebles et~al.(2022)Peebles, Zhu, Zhang, Torralba, Efros, and
  Shechtman}]{peebles2022gansupervised}
Peebles W, Zhu JY, Zhang R, Torralba A, Efros A, Shechtman E (2022)
  Gan-supervised dense visual alignment. In: CVPR

\bibitem[{Perez et~al.(2018)Perez, Strub, De~Vries, Dumoulin, and
  Courville}]{perez2018film}
Perez E, Strub F, De~Vries H, Dumoulin V, Courville A (2018) Film: Visual
  reasoning with a general conditioning layer. In: AAAI, vol~32

\bibitem[{Pumarola et~al.(2020)Pumarola, Corona, Pons-Moll, and
  Moreno-Noguer}]{pumarola_d-nerf_2020}
Pumarola A, Corona E, Pons-Moll G, Moreno-Noguer F (2020) {D-NeRF: Neural
  Radiance Fields for Dynamic Scenes}. In: CVPR

\bibitem[{Qi et~al.(2017)Qi, Su, Mo, and Guibas}]{Qi2017PointNetDL}
Qi C, Su H, Mo K, Guibas L (2017) Pointnet: Deep learning on point sets for 3d
  classification and segmentation. In: CVPR, pp 77--85

\bibitem[{Richardson et~al.(2021)Richardson, Alaluf, Patashnik, Nitzan, Azar,
  Shapiro, and Cohen-Or}]{richardson2020encoding}
Richardson E, Alaluf Y, Patashnik O, Nitzan Y, Azar Y, Shapiro S, Cohen-Or D
  (2021) Encoding in style: a stylegan encoder for image-to-image translation.
  In: CVPR

\bibitem[{Sahillioglu(2019)}]{Sahillioglu2019RecentAI}
Sahillioglu Y (2019) Recent advances in shape correspondence. The Visual
  Computer 36:1705 -- 1721

\bibitem[{Schwarz et~al.(2020)Schwarz, Liao, Niemeyer, and
  Geiger}]{Schwarz2020NEURIPS}
Schwarz K, Liao Y, Niemeyer M, Geiger A (2020) Graf: Generative radiance fields
  for 3d-aware image synthesis. In: NIPS

\bibitem[{Shen et~al.(2020)Shen, Yang, Tang, and Zhou}]{Shen2020InterFaceGANIT}
Shen Y, Yang C, Tang X, Zhou B (2020) Interfacegan: Interpreting the
  disentangled face representation learned by gans. PAMI PP

\bibitem[{Simonyan and Zisserman(2015)}]{Simonyan2015VeryDC}
Simonyan K, Zisserman A (2015) Very deep convolutional networks for large-scale
  image recognition. In: CoRR, vol abs/1409.1556

\bibitem[{Sitzmann et~al.(2020)Sitzmann, Martel, Bergman, Lindell, and
  Wetzstein}]{sitzmann2020siren}
Sitzmann V, Martel JN, Bergman AW, Lindell DB, Wetzstein G (2020) Implicit
  neural representations with periodic activation functions. In: NIPS

\bibitem[{Tancik et~al.(2020)Tancik, Srinivasan, Mildenhall, Fridovich-Keil,
  Raghavan, Singhal, Ramamoorthi, Barron, and Ng}]{tancik_fourier_2020}
Tancik M, Srinivasan PP, Mildenhall B, Fridovich-Keil S, Raghavan N, Singhal U,
  Ramamoorthi R, Barron JT, Ng R (2020) Fourier features let networks learn
  high frequency functions in low dimensional domains. In: NIPS

\bibitem[{Tewari et~al.(2020)Tewari, Fried, Thies, Sitzmann, Lombardi,
  Sunkavalli, Martin-Brualla, Simon, Saragih, Nie{\ss}ner, Pandey, Fanello,
  Wetzstein, Zhu, Theobalt, Agrawala, Shechtman, Goldman, and
  Zollh{\"o}fer}]{Tewari2020NeuralSTAR}
Tewari A, Fried O, Thies J, Sitzmann V, Lombardi S, Sunkavalli K,
  Martin-Brualla R, Simon T, Saragih J, Nie{\ss}ner M, Pandey R, Fanello S,
  Wetzstein G, Zhu JY, Theobalt C, Agrawala M, Shechtman E, Goldman DB,
  Zollh{\"o}fer M (2020) {State of the Art on Neural Rendering}. Computer
  Graphics Forum

\bibitem[{Tov et~al.(2021)Tov, Alaluf, Nitzan, Patashnik, and
  Cohen-Or}]{tov2021designing}
Tov O, Alaluf Y, Nitzan Y, Patashnik O, Cohen-Or D (2021) Designing an encoder
  for stylegan image manipulation. arXiv

\bibitem[{Tritrong et~al.(2021)Tritrong, Rewatbowornwong, and
  Suwajanakorn}]{Tritrong2021RepurposeGANs}
Tritrong N, Rewatbowornwong P, Suwajanakorn S (2021) Repurposing gans for
  one-shot semantic part segmentation. In: CVPR

\bibitem[{Truong et~al.(2020{\natexlab{a}})Truong, Danelljan, Gool, and
  Timofte}]{Truong2020GOCorBG}
Truong P, Danelljan M, Gool LV, Timofte R (2020{\natexlab{a}}) {GOCor}:
  Bringing globally optimized correspondence volumes into your neural network.
  In: NIPS

\bibitem[{Truong et~al.(2020{\natexlab{b}})Truong, Danelljan, and
  Timofte}]{Truong2020GLUNetGU}
Truong P, Danelljan M, Timofte R (2020{\natexlab{b}}) Glu-net: Global-local
  universal network for dense flow and correspondences. In: CVPR, pp 6257--6267

\bibitem[{Truong et~al.(2021)Truong, Danelljan, Gool, and
  Timofte}]{Truong2021LearningAD}
Truong P, Danelljan M, Gool LV, Timofte R (2021) Learning accurate dense
  correspondences and when to trust them. In: CVPR, pp 5710--5720

\bibitem[{Wang et~al.(2019{\natexlab{a}})Wang, Bo, and Fuxin}]{Wang_2019_ICCV}
Wang X, Bo L, Fuxin L (2019{\natexlab{a}}) Adaptive wing loss for robust face
  alignment via heatmap regression. In: ICCV

\bibitem[{Wang et~al.(2019{\natexlab{b}})Wang, Sun, Liu, Sarma, Bronstein, and
  Solomon}]{Wang2019DynamicGC}
Wang Y, Sun Y, Liu Z, Sarma SE, Bronstein M, Solomon J (2019{\natexlab{b}})
  Dynamic graph cnn for learning on point clouds. TOG 38:1 -- 12

\bibitem[{Wang et~al.(2021)Wang, Bagautdinov, Lombardi, Simon, Saragih,
  Hodgins, and Zollhofer}]{Wang_2021_CVPR}
Wang Z, Bagautdinov T, Lombardi S, Simon T, Saragih J, Hodgins J, Zollhofer M
  (2021) Learning compositional radiance fields of dynamic human heads. In:
  CVPR, pp 5704--5713

\bibitem[{Xu and Wang(2021)}]{xu2021rethinking}
Xu J, Wang X (2021) Rethinking self-supervised correspondence learning: A video
  frame-level similarity perspective. arXiv

\bibitem[{Yang et~al.(2021)Yang, Belongie, Hariharan, and
  Koltun}]{yang2021geometry}
Yang G, Belongie S, Hariharan B, Koltun V (2021) Geometry processing with
  neural fields. In: Thirty-Fifth Conference on Neural Information Processing
  Systems

\bibitem[{Yang et~al.(2022)Yang, Jiang, Liu, , and Loy}]{yang2022Unsupervised}
Yang S, Jiang L, Liu Z, , Loy CC (2022) Unsupervised image-to-image translation
  with generative prior. In: CVPR

\bibitem[{Yu et~al.(2021)Yu, Ye, Tancik, and Kanazawa}]{yu2021pixelnerf}
Yu A, Ye V, Tancik M, Kanazawa A (2021) pixelnerf: Neural radiance fields from
  one or few images. In: CVPR, pp 4578--4587

\bibitem[{Zhang et~al.(2016)Zhang, Zhang, Li, and Qiao}]{Zhang2016JointFD}
Zhang K, Zhang Z, Li Z, Qiao Y (2016) Joint face detection and alignment using
  multitask cascaded convolutional networks. IEEE Signal Processing Letters
  23:1499--1503

\bibitem[{Zhang et~al.(2020)Zhang, Riegler, Snavely, and
  Koltun}]{zhang2020nerf++}
Zhang K, Riegler G, Snavely N, Koltun V (2020) Nerf++: Analyzing and improving
  neural radiance fields. arXiv

\bibitem[{Zhang et~al.(2018)Zhang, Isola, Efros, Shechtman, and
  Wang}]{zhang2018perceptual}
Zhang R, Isola P, Efros AA, Shechtman E, Wang O (2018) The unreasonable
  effectiveness of deep features as a perceptual metric. In: CVPR

\bibitem[{Zhang et~al.(2008)Zhang, Sun, and Tang}]{cats}
Zhang W, Sun J, Tang X (2008) Cat head detection - how to effectively exploit
  shape and texture features. In: ECCV

\bibitem[{Zhang et~al.(2021{\natexlab{a}})Zhang, Chen, Ling, Gao, Zhang,
  Torralba, and Fidler}]{StyleGAN3D}
Zhang Y, Chen W, Ling H, Gao J, Zhang Y, Torralba A, Fidler S
  (2021{\natexlab{a}}) Image gans meet differentiable rendering for inverse
  graphics and interpretable 3d neural rendering. In: ICLR

\bibitem[{Zhang et~al.(2021{\natexlab{b}})Zhang, Ling, Gao, Yin, Lafleche,
  Barriuso, Torralba, and Fidler}]{Zhang2021DatasetGANEL}
Zhang Y, Ling H, Gao J, Yin K, Lafleche JF, Barriuso A, Torralba A, Fidler S
  (2021{\natexlab{b}}) Datasetgan: Efficient labeled data factory with minimal
  human effort. In: CVPR

\bibitem[{Zheng et~al.(2022)Zheng, Abrevaya, B{\"u}hler, Chen, Black, and
  Hilliges}]{zheng2022imavatar}
Zheng Y, Abrevaya VF, B{\"u}hler MC, Chen X, Black MJ, Hilliges O (2022) {I}
  {M} {Avatar}: Implicit morphable head avatars from videos. In: CVPR

\bibitem[{Zheng et~al.(2021)Zheng, Yu, Dai, and Liu}]{zheng2021deep}
Zheng Z, Yu T, Dai Q, Liu Y (2021) Deep implicit templates for 3d shape
  representation. In: CVPR, pp 1429--1439

\bibitem[{Zhou et~al.(2021)Zhou, Xie, Ni, and Tian}]{zhou2021CIPS3D}
Zhou P, Xie L, Ni B, Tian Q (2021) {{CIPS}}-{{3D}}: A {{3D}}-{{Aware
  Generator}} of {{GANs Based}} on {{Conditionally}}-{{Independent Pixel
  Synthesis}}. arXiv \eprint{2110.09788}

\bibitem[{Zhou et~al.(2016)Zhou, Kr{\"a}henb{\"u}hl, Aubry, Huang, and
  Efros}]{Zhou2016LearningDC}
Zhou T, Kr{\"a}henb{\"u}hl P, Aubry M, Huang Q, Efros AA (2016) Learning dense
  correspondence via 3d-guided cycle consistency. In: CVPR, pp 117--126

\end{thebibliography}

    \end{document}